\documentclass[11pt]{article}

\PassOptionsToPackage{table}{xcolor}

\usepackage[preprint]{acl}

\usepackage{times}
\usepackage{latexsym}
\usepackage[T1]{fontenc}
\usepackage[utf8]{inputenc}
\usepackage{graphicx}
\usepackage{amsmath}
\usepackage{amssymb}
\usepackage{amsthm}
\usepackage{booktabs}
\usepackage{multirow}
\usepackage{enumitem}

\usepackage{algorithm}
\usepackage{algpseudocode}

\usepackage{tikz}
\usetikzlibrary{positioning, shapes.geometric, arrows.meta, calc}
\usepackage{pgfplots}
\pgfplotsset{compat=1.18}

\usepackage{float}
\usepackage[most,skins,breakable]{tcolorbox}
\definecolor{promptaccent}{HTML}{34708F}
\definecolor{prompttint}{HTML}{F2F6F8}
\newtcolorbox{promptbox}[1]{%
  enhanced, breakable,
  colback=prompttint, colframe=promptaccent, boxrule=0.5pt,
  arc=1.5pt, outer arc=1.5pt,
  left=5pt, right=5pt, bottom=4pt, top=2pt,
  coltitle=white, fonttitle=\bfseries\footnotesize, colbacktitle=promptaccent,
  toptitle=2pt, bottomtitle=2pt, lefttitle=6pt, righttitle=6pt,
  title={#1}, before skip=4pt, after skip=5pt, fontupper=\footnotesize}
\newcommand{\promptlabel}[1]{\par\vskip2pt\noindent\textbf{\footnotesize #1}\nobreak\par\nobreak\vskip1pt\nobreak}
\newcommand{\promptjson}[1]{{\par\ttfamily\footnotesize\raggedright #1\par}}

\theoremstyle{definition}

\newtheorem{definition}{Definition}

\title{To Isolate or to Score? Model-Adaptive Assessment \\ for Cost-Efficient Multi-Agent RAG}

\author{
  \bfseries Jungseob Lee$^{1}$ \quad
  Chanjun Park$^{2}$\thanks{Corresponding authors.} \quad
  Heuiseok Lim$^{1}$\footnotemark[1] \\
  $^{1}$Korea University \quad $^{2}$Soongsil University \\
  \texttt{\{omanma1928, limhseok\}@korea.ac.kr} \\
  \texttt{chanjun.park@ssu.ac.kr}
}

\begin{document}
\maketitle
\begin{abstract}
Multi-agent document assessment for retrieval-augmented generation is computationally expensive, driving practitioners toward smaller, deployable models whose assessment mechanisms remain poorly understood. We conduct a controlled study of training-free interventions on 7B--9B instruction-tuned models across diverse QA benchmarks, revealing a sharp dichotomy in how models benefit from assessment. For weaker baselines, the dominant mechanism is per-document isolation. Astoundingly, assessment-free isolation matches full multi-agent assessment, demonstrating that resolving multi-document context confusion, rather than scoring quality, drives outsized gains of up to 50 percentage points. Conversely, for strong baselines where scoring quality matters, we introduce Reasoning-Score Coupling, a label-free perturbation probe that classifies scoring behavior. Integrating these findings, we propose MADARA, a model-adaptive routing architecture. Crucially, MADARA's diagnostic thresholds derived from a single pilot model generalize zero-shot to four unseen model families, providing a robust, lightweight pipeline to eliminate computational overhead.
\end{abstract}

\section{Introduction}
\label{sec:intro}

Multi-agent document assessment is increasingly used to improve retrieval-augmented generation (RAG) by deploying specialized agents to evaluate, filter, or debate retrieved documents~\citep{lewis2020retrieval, chang2025main, wang2025retrieval, hu2025removal}. Despite its popularity, this paradigm multiplies inference calls by \(\mathcal{O}(T_{\text{rounds}} \times N_{\text{agents}} \times N_{\text{docs}})\) compared to standard RAG, especially when iterative consensus or multi-perspective scoring is required. In production environments, applying this combinatorial multiplier to massive (30B+) language models results in prohibitive latency and financial costs. To circumvent this overhead, practitioners are increasingly turning to smaller, more cost-effective models (e.g., 7B--9B parameters)~\cite{wang2025comprehensive, lu2024small, cai2026efficient, prieto2025edge}.

However, this pragmatic shift creates a critical mechanistic misalignment: do these smaller, deployable models actually possess the sophisticated reasoning capabilities required to conduct meaningful multi-agent assessment? If improvements primarily stem from structural side-effects (e.g., bypassing long-context confusion) rather than from the agents' substantive reasoning, the community may be paying massive compute costs for theoretically redundant processing~\cite{liu2024lost}.

Indeed, our initial probing of these deployable models reveals a highly inconsistent landscape. We observe that the exact same assessment pipeline can boost exact-match accuracy by more than ten percentage points for one model while actively degrading it for another~\cite{luo2025zero, gao2026u, du2025context}. This stark contrast raises a fundamental mechanistic question: what drives the gains when they occur, and why do they fail to transfer across models?

To bridge this gap, we address this directly through controlled ablations of a standard three-agent RAG pipeline~\citep{chang2025main}. Using only training-free interventions (prompting, aggregation, and generation strategies), we analyze the behavior of instruction-tuned models across diverse QA benchmarks. Building on these mechanistic insights, we propose Model-Adaptive Document Assessment Routing Architecture (MADARA), a dynamic pipeline that automatically routes model--task pairs to their optimal, most cost-effective assessment strategy.

Our central claim is that \textbf{assessment value is gated by an intrinsic capability divide that depends on the interaction of model capacity with task structure}. We operationalise this claim as a \textbf{Diagnose$\to$Treat} pipeline: RSC together with the No-Filter baseline diagnoses the model--task pair, and the appropriate treatment among PDE, SDA, CoT, and ATF follows directly from that diagnosis. The two findings below are the diagnosis and treatment halves of this single claim, not independent contributions.

\noindent\textbf{1. Diagnosis: a sharp isolation--scoring asymmetry governs which assessment regime applies.} We identify the asymmetry empirically. For weaker models, \emph{per-document isolation} drives outsized gains, boosting performance by 25 to 36 percentage points on adversarial conflicts and by up to 50 percentage points on standard QA. Astoundingly, even random, assessment-free isolation matches full multi-agent variants. This proves that resolving multi-document context confusion, rather than scoring quality, is the actual bottleneck, rendering heavy multi-agent compute redundant and reducing inference calls by roughly \(4\times\). Conversely, strong-baseline models show no benefit from isolation. Therefore, Per-Document Extraction (PDE) can retain peak performance while entirely eliminating assessment overhead in the weak-baseline regime.

\noindent\textbf{2. Treatment: RSC-driven MADARA routing transfers zero-shot.} For strong models, scoring quality remains critical. We introduce \emph{Reasoning-Score Coupling} (RSC), a perturbation-based probe~\citep{lanham2023measuring, paul2024making} that classifies model--task pairs as \emph{quality-ordered} or \emph{stochastic} using only 100 unlabeled queries; RSC is the diagnostic arm. MADARA integrates RSC with the No-Filter baseline to route each model--task pair to its optimal treatment; MADARA is the treatment arm. Crucially, routing thresholds derived from a single pilot model transfer zero-shot to four unseen model families. This confirms the capability divide is intrinsic rather than an overfitted artifact, providing a robust, lightweight pipeline to eliminate computational waste. Furthermore, our mechanistic findings regarding the necessity of isolation persist even when upgrading from sparse to state-of-the-art dense retrieval and generative reranking.

\section{Background and Related Work}
\label{sec:related}

\paragraph{Multi-Agent Debate and Assessment.}
Multi-agent debate for improving LLM factuality~\citep{du2024improving} has expanded into broader agentic RAG architectures~\citep{singh2025agentic}. Subsequent works address sycophantic convergence~\citep{liang2024encouraging, jain2025beyond, pitre2025consensagent, zhu2026demystifying}, majority voting~\citep{smit2023should}, consensus-free alternatives~\citep{cui2025free}, and mental-set diversification~\citep{liu2025breaking}. In RAG, MADAM-RAG~\citep{wang2025retrieval} debates answers using 70B+ agents, while DRAG~\citep{hu2025removal}, MA-RAG~\citep{nguyen2025ma}, and MAIN-RAG~\citep{chang2025main} explore reranking and filtering. Astute-RAG~\citep{wang2025astute} consolidates knowledge internally within a single model; architecturally, our setting differs by studying how \emph{multiple assessment agents} interact with a downstream generator, though our per-document extraction (PDE) finding could complement Astute-RAG when internal consolidation is insufficient. Crucially, while existing literature largely treats multi-agent assessment as a black-box performance enhancer, our controlled PDE-Random ablation (Table~\ref{tab:pde-ablation}) specifically isolates this mechanism. This provides a \emph{mechanistic proof} that the entire gain for weak models comes strictly from isolation, rendering assessment compute irrelevant, a diagnostic claim fundamentally addressing the gap in mechanistic understanding.

\paragraph{Adaptive and Self-Correcting RAG.}
Adaptive systems route queries based on per-query signals: learned reflection tokens~\citep{asai2024self}, corrective search~\citep{yan2024corrective}, complexity classifiers~\citep{jeong2024adaptive}, or preference data~\citep{ong2024routellm}. Concurrent works further explore cooperative RL optimization~\citep{chen2025improving}, knowledge-graph conflict resolution~\citep{liu2025truthfulrag}, search conflict detection~\citep{cattan2025dragged}, information-gain reranking~\citep{wang2025infogain}, and conflict taxonomies~\citep{xu2024knowledge}. In contrast, our RSC-based routing operates at the \emph{model--domain level} (a one-time probe applied uniformly to all queries), requiring no training and only 100 queries.

\paragraph{Score Calibration and CoT Faithfulness.}
LLM score calibration is widely studied in the judge setting~\citep{jung2024trust, jain2025beyond, li2025cot, pitre2025consensagent, li2025generation}. More directly relevant is the CoT faithfulness literature, showing LLMs often produce unfaithful explanations~\citep{turpin2023language, lanham2023measuring}. Subsequent work quantifies this via causal mediation~\citep{paul2024making}, examines reasoning--answer correlations~\citep{jiang2025makes}, unlearning~\citep{tutek2025measuring}, and fundamental limits~\citep{lyu2023faithful, tanneru2024hardness}. Concurrently, MATCHA~\citep{jiang2025robust} probes whether CoT answers decouple from reasoning under perturbation. Our RSC diagnostic adapts this to target multi-agent \emph{scoring}. By testing whether numerical scores degrade monotonically as reasoning quality declines, RSC provides a fine-grained diagnostic to measure reasoning--score coupling strength, allowing us to prescribe targeted remedies like CoT de-polarization for uncalibrated models. (Extended related work is provided in Appendix~\ref{sec:appendix-related}).

\section{Reasoning-Score Coupling}
\label{sec:rsc}

\paragraph{The diagnostic arm of Diagnose$\to$Treat.}
RSC is the \emph{diagnostic} half of the pipeline introduced in \S\ref{sec:intro}: it identifies, without gold labels, whether a model's scoring behaviour on a given task is reasoning-driven or stochastic, and the corresponding treatment (\S\ref{sec:madara}) follows directly from the diagnosis. We introduce \emph{Reasoning-Score Coupling} (RSC), a task-specific diagnostic detecting whether a model's document scores degrade monotonically under systematic reasoning perturbation. If scores track reasoning quality, degrading reasoning monotonically decreases score reliability; a lack of this pattern indicates CoT interventions are unlikely to help. RSC provides \emph{correlational} evidence of a reasoning--scoring association; establishing causality requires intervention-based methods \citep{paul2024making}. The probe measures \emph{scoring sensitivity to reasoning perturbation}, distinct from standard CoT faithfulness (Appendix~\ref{sec:appendix-rsc-details}).

\subsection{Perturbation Protocol}
\label{subsec:rsc-protocol}

RSC compares a model's document scores under CoT reasoning against those obtained after three levels of increasing perturbation: \textbf{(1) Shuffled} (reasoning steps randomly reordered; disrupts logical flow but preserves semantics), \textbf{(2) Contradicted} (steps semantically negated; disrupts both coherence and directional cues), and \textbf{(3) Random} (reasoning from a completely different query-document pair; entirely irrelevant). 

For each level \(k \in \{1, 2, 3\}\), we compute the Spearman rank correlation \(\hat{\rho}_k\) between normal (\(\mathbf{s}_{\text{normal}} \in [0,5]^m\)) and perturbed scores (\(\mathbf{s}_{P_k}\)) across all calibration documents:
\begin{equation}
  \hat{\rho}_k = \text{Spearman}\bigl(\mathbf{s}_{\text{normal}},\,\mathbf{s}_{P_k}\bigr)
  \label{eq:rho-k}
\end{equation}
The severity ordering is motivated \emph{a priori}, representing monotonically decreasing preserved information (empirical confirmation in Appendix~\ref{sec:appendix-rsc-details}).

\subsection{Formal Definition of RSC}
\label{subsec:rsc-formal}

\noindent\textbf{Terminology.} We use ``perturbation-level correlations \(\hat{\rho}_k\)'' for Equation~\ref{eq:rho-k} and ``trend coefficient \(\rho^*\)'' for the final monotonicity statistic.

\begin{definition}[Reasoning-Score Coupling, RSC]
Let \(M\) be a language model and \(\mathcal{D}\) a calibration set of \(n\) examples each with \(d_i\) retrieved documents. The \emph{RSC trend coefficient} is:
\begin{equation}
  \rho^* = \text{Spearman}\bigl([1,\,2,\,3],\;[\hat{\rho}_1,\,\hat{\rho}_2,\,\hat{\rho}_3]\bigr)
  \label{eq:trend-rho}
\end{equation}
Model \(M\) on domain \(\mathcal{D}\) is classified as:
\begin{align*}
  \text{quality-ordered} &\quad \text{if } \rho^* = -1.0 \\
  \text{stochastic (non-monotonic)} &\quad \text{otherwise}
\end{align*}
Since \(\rho^*\) takes only five discrete values, this reduces to a deterministic check for perfect monotonic degradation (\(\hat{\rho}_1 > \hat{\rho}_2 > \hat{\rho}_3\)). The metric's reliability derives from the highly significant per-level \(\hat{\rho}_k\) values (\(p < 0.001\)) and bootstrap resampling (Appendix~\ref{sec:appendix-split-half}, Table~\ref{tab:bootstrap-rho}).
\end{definition}

\noindent Beyond the binary classification, \(\hat{\rho}_1\) measures the \emph{strength} of baseline coupling, defining three states: \emph{stochastic} (\(\rho^* > -1.0\)), \emph{weakly coupled} (\(\rho^* = -1.0\), \(\hat{\rho}_1 < 0.5\)), and \emph{strongly coupled} (\(\rho^* = -1.0\), \(\hat{\rho}_1 \geq 0.5\)). The protocol requires no gold labels: the quality oracle is the model's own generated reasoning. Beyond this binary trend, we additionally report a continuous, magnitude-aware companion signal \(\bar{\rho} = \tfrac{1}{3}(\hat{\rho}_1 + \hat{\rho}_2 + \hat{\rho}_3)\), used as an explicit robustness check (Appendix~\ref{sec:appendix-rho-bar}). Furthermore, RSC captures process-level coupling rather than distributional spread, making it a distinct and more robust routing signal than standard score entropy (see Appendix~\ref{sec:appendix-rsc-entropy}).

\begin{figure*}[t]
\centering
\includegraphics[width=0.99\linewidth]{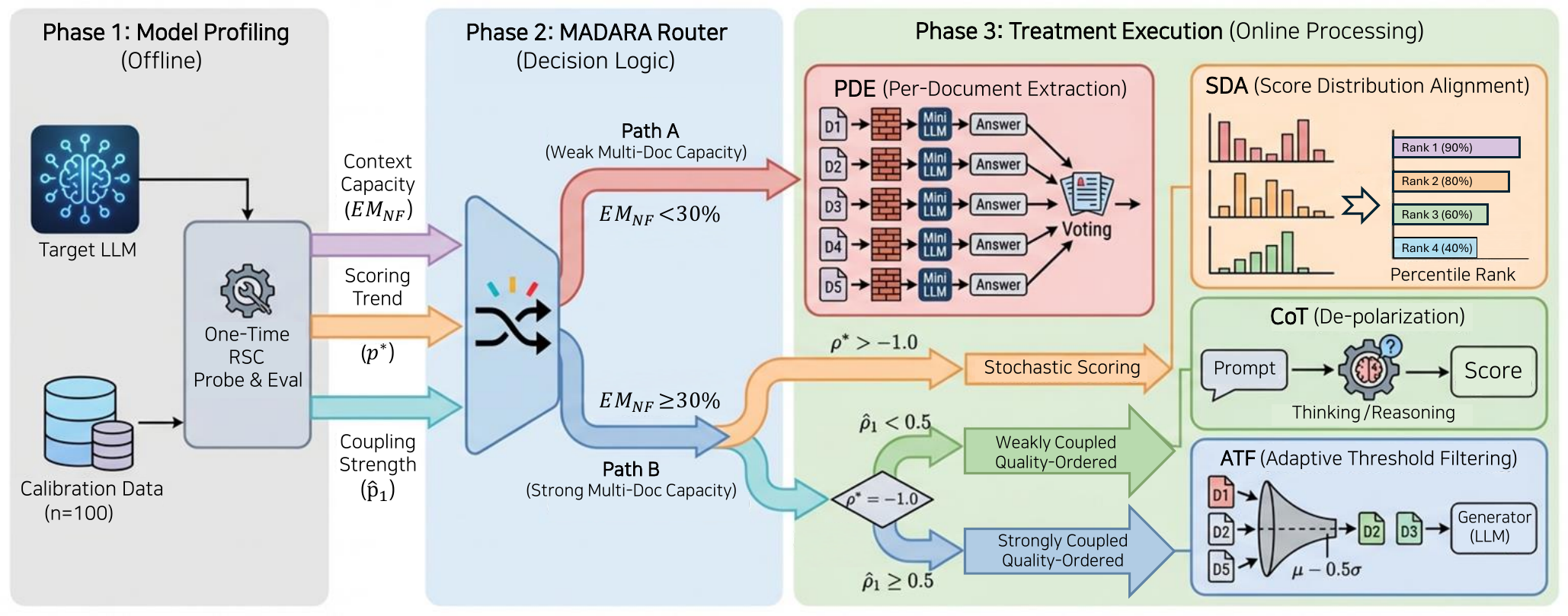}
\caption{The MADARA Model-Adaptive Routing Architecture. (Left) A one-time RSC probe evaluates the target LLM's context capacity (\(\text{EM}_{\text{NF}}\)) and scoring behavior (\(\rho^*, \hat{\rho}_1\)). (Middle) The router identifies a capability phase-transition: weak-baseline models are strictly routed to bypass multi-document evaluation. (Right) PDE (Isolation) structurally separates documents to cure context confusion, while scoring-only treatments (SDA, CoT, ATF) refine document assessment for strong models. Full pseudocode is provided in Appendix~\ref{sec:appendix-madara-algorithm}.}
\label{fig:madara-overview}
\end{figure*}

\section{Candidate Treatment Strategies}
\label{sec:madara}

\paragraph{The treatment arm of Diagnose$\to$Treat.}
MADARA is the \emph{treatment} half of the pipeline. The four candidate strategies below address distinct failure modes diagnosed by RSC (\S\ref{sec:rsc}) and the No-Filter (NF) baseline; the routing decisions are operational consequences of the capability divide identified by these two diagnostic signals, not independent design choices. We evaluate the four strategies on real and synthetic failure modes, and route a model--task pair to its corresponding treatment based on two diagnostics: RSC (scoring behaviour) and NF accuracy (context-handling capacity). All treatments operate within a three-agent assessment framework (agent design and prompts in Appendix~\ref{sec:appendix-agent-prompts}).

\subsection{CoT De-Polarization}
\label{subsec:cot-depol}

For quality-ordered models (\(\rho^* = -1.0\)), the baseline failure mode is \emph{polarization}: without reasoning guidance, agents assign extreme scores (0 or 5) to \({\approx}80\%\) of documents. CoT de-polarization requires agents to generate explicit reasoning before scoring, mitigating this direct-to-extreme pattern. For example, extreme scores for Mistral-7B on CONFLICTS drop from \(80.8\%\) to \(3.5\%\), yielding a \(+4.7\)pp EM gain over NF (Table~\ref{tab:main-results}).

\subsection{Score Distribution Alignment}
\label{subsec:sda}

For stochastic models (\(\rho^* > -1.0\)), reasoning quality does not drive scores, rendering CoT de-polarization ineffective. \emph{Score Distribution Alignment} (SDA) bypasses reasoning entirely. It converts each agent's raw scores to percentile ranks, then aggregates them via weighted averaging to produce a calibrated ranking (Algorithm~\ref{alg:sda}, Appendix~\ref{sec:appendix-method-details}).

\subsection{Adaptive Threshold Filtering}
\label{subsec:aga-thresh}

For strongly coupled models (\(\rho^* = -1.0\), \(\hat{\rho}_1 \geq 0.5\)), baselines already produce effective rankings. ATF leverages these scores to \emph{filter} rather than rerank:
\begin{equation}
  \tau = \mu(\mathbf{s}) - \kappa \cdot \sigma(\mathbf{s})
\end{equation}
(with \(\kappa = 0.5\)). It retains only documents with \(s_i \geq \tau\) (minimum 2, maximum \(k\)), requiring no additional LLM calls.

\subsection{Per-Document Answer Extraction}
\label{subsec:aga-perdoc}

For models struggling with multi-document context (low NF accuracy), \emph{Per-Document Extraction} (PDE) structurally decomposes generation: (1)~3-agent scoring evaluates documents; (2)~the model generates an answer from each top-\(k\) document individually; (3)~candidates are grouped by normalized string match, and the group with the highest cumulative score is selected. Component ablations (\S~\ref{sec:results}) reveal that this isolation, rather than scoring quality, drives PDE's outsized gains.

\subsection{MADARA Routing Protocol}
\label{subsec:madara-pipeline}

The MADARA pipeline (Figure~\ref{fig:madara-overview}) operates in two phases. Phase 1 runs a one-time RSC probe to classify the model-task pair. Phase 2 processes each query using the selected optimal treatment: SDA for stochastic models; PDE for quality-ordered models with weak NF baselines; and CoT de-polarization (\(\hat{\rho}_1 < 0.5\)) or ATF (\(\hat{\rho}_1 \geq 0.5\)) for quality-ordered models with strong NF baselines. The NF threshold is estimated from the RSC calibration set (pseudocode in Appendix~\ref{sec:appendix-madara-algorithm}).

\section{Experimental Setup}
\label{sec:experiments}

\subsection{Models}
\label{subsec:models}

To investigate the target regime of cost-effective, deployable LLMs, we evaluate five open-weight, instruction-tuned 7B--9B models:
\textit{Llama-3.1-8B-Instruct} \citep{grattafiori2024llama},
\textit{Mistral-7B-Instruct-v0.3} \citep{jiang2023mistral7b},
\textit{Qwen3-8B} \citep{yang2025qwen3},
\textit{Qwen2.5-7B-Instruct} \citep{qwen2025qwen25technicalreport},
and \textit{Gemma-2-9B-IT} \citep{team2024gemma}.
All models are served in \texttt{bfloat16} via vLLM~v0.8.5 \citep{kwon2023efficient} with \texttt{max\_seq\_len}\(=\)4096 on NVIDIA A100/RTX8000 GPUs at temperature \(0.6\).

\subsection{Benchmarks}
\label{subsec:benchmarks}

To manage the massive \(\mathcal{O}(T \times N \times D)\) multi-agent inference overhead, we evaluate on sampled subsets (\({\sim}\)1K queries) of three diverse datasets: \textbf{(1) CONFLICTS (CFL)} \citep{xie2023adaptive}: an adversarial QA benchmark with inter-document contradictions via entity substitution, filtered for strict exact-match fidelity. \textbf{(2) FEVER (FVR)} \citep{thorne2018fever}: a binary fact-verification task with BM25 retrieval, representing a high-baseline regime. \textbf{(3) TriviaQA (TQA)} \citep{joshi2017triviaqa}: a standard factoid QA benchmark with BM25 retrieval, serving as our held-out set to replicate the isolation and RSC findings (\S~\ref{sec:results}).

\subsection{Methods Compared}
\label{subsec:methods}

We compare six strategies across two categories. \emph{Scoring treatments}: \textbf{(1) NF}: all documents to generator (standard RAG); \textbf{(2) 3-Agent Baseline}: weighted score aggregation (0.4, 0.3, 0.3); \textbf{(3) CoT De-Polarization}: explicit reasoning before scoring (\S~\ref{subsec:cot-depol}); \textbf{(4) SDA}: percentile-rank normalization (SDA outperforms standard RRF by adapting to heterogeneous distributions; see Appendix~\ref{sec:appendix-sda-ablation}). \emph{Generation strategies}: \textbf{(5) ATF}: adaptive threshold (\(\mu - 0.5\sigma\)) filtering (\S~\ref{subsec:aga-thresh}); \textbf{(6) PDE}: per-document extraction with score-weighted majority voting (\S~\ref{subsec:aga-perdoc}). The MADARA router dynamically selects among these treatments. 

We exclude MADAM-RAG \citep{wang2025retrieval} (which debates generated answers using 70B+ models) and FiD \citep{izacard2021leveraging} (which modifies cross-attention) to strictly isolate training-free, document-level assessment behaviors in the 7B--9B regime. A detailed discussion on baseline scope is provided in Appendix~\ref{sec:appendix-baselines-scope}.

\subsection{Evaluation Metrics}
\label{subsec:metrics}

We report Exact Match (EM; strictly matched after whitespace normalization and lowercasing) and Token F1 (harmonic mean of token-level precision and recall). For the binary FEVER task, EM and F1 are equivalent.

\subsection{Implementation Details}
\label{subsec:implementation}

Agents share the same base model per experiment, reranking top-\(k{=}5\) from 10 retrieved documents. The RSC probe uses \(n{=}100\) label-free calibration queries. SDA uses uniform agent weights \(\mathbf{w} = (1/3, 1/3, 1/3)\). Crucially, routing thresholds (\(\rho^* = -1.0\), \(\hat{\rho}_1 = 0.5\), \(\tau_{\text{NF}} = 30\%\)) were derived strictly from a single pilot model (Mistral-7B) to prevent overfitting and applied zero-shot to all others (sensitivity analyses in Appendices~\ref{sec:appendix-rho1-sensitivity} and~\ref{sec:appendix-tau-sensitivity}).

\section{Results}
\label{sec:results}

\subsection{Isolation vs.\ Scoring}
\label{subsec:finding-isolation}

\paragraph{Isolation dominates for weak models.}
PDE provides outsized gains exclusively for weak-baseline models (Table~\ref{tab:main-results}), yielding \(+36.3\)pp for Llama and \(+25.4\)pp for Mistral on CONFLICTS (\(p < 0.001\)). In contrast, scoring-only treatments provide at most \(+5.5\)pp across all models. Crucially, this isolation finding replicates on a held-out factoid benchmark (TriviaQA, Table~\ref{tab:pde-ablation}), where Llama gains \(+49.8\)pp. This provides strong confirmation that the outsized PDE gain is a fundamental mechanism, not an artifact of CONFLICTS' adversarial document structure.

\paragraph{Assessment quality is redundant for weak baselines.}
A component ablation (Table~\ref{tab:pde-ablation}) confirms that structural isolation, rather than assessment quality, drives these gains. PDE-Random (random document selection with uniform voting) completely bypasses multi-agent assessment yet matches the full PDE pipeline for Llama on both CONFLICTS (\(50.6\%\) vs.\ \(50.2\%\)) and TriviaQA (\(79.6\%\) vs.\ \(79.6\%\)). For Mistral, assessment-guided selection adds \(+19\)pp beyond random isolation on CONFLICTS. By bypassing multi-agent evaluation, PDE-Random reduces inference calls by roughly \(4\times\). This proves that resolving context confusion renders heavy assessment compute largely redundant for weak baselines. Token F1 scores further verify that these improvements are not mere formatting artifacts (see Appendix~\ref{sec:appendix-f1-mechanistic}).

\begin{table}[t]
\centering
\resizebox{\columnwidth}{!}{%
\begin{tabular}{@{}ll c ccc@{}}
\toprule
& & \textbf{Base} & \multicolumn{3}{c}{\textbf{PDE Component Ablation}} \\
\cmidrule(l){4-6}
\textbf{Model} & \textbf{Task} & \textbf{NF} & \textbf{Rand.} & \textbf{Unif.} & \textbf{Full} \\
& & \scriptsize{(\(1\times\) cost)} & \scriptsize{(\(1\times\) cost)} & \scriptsize{(\(\approx 4\times\))} & \scriptsize{(\(\approx 4\times\))} \\
\midrule
Llama-3.1 & Conflicts & 13.9 & \textbf{50.6} & \textbf{51.1} & \textbf{50.2} \\
 & TriviaQA & 29.8 & \textbf{79.6} & \textbf{80.1} & \textbf{79.6} \\
\midrule
Mistral-v0.3 & Conflicts & 18.1 & 24.5 & 38.8 & \textbf{43.5} \\
 & TriviaQA & 67.6 & 71.5 & \textbf{73.7} & \textbf{73.7} \\
\bottomrule
\end{tabular}%
}
\caption{\textbf{PDE component ablation (EM\%) reveals that multi-agent assessment is redundant for weak baselines.} For Llama, completely assessment-free isolation (\textit{Rand.}) yields identical outsized gains as the computationally heavy \textit{Full} pipeline, proving that resolving context confusion drives the improvement. (\textit{Cost multipliers} indicate relative inference calls vs.\ standard RAG. CFL=Conflicts, TQA=TriviaQA; NF=No Filter; Unif.=assessment-guided + uniform vote; Full=score-weighted vote.)}
\label{tab:pde-ablation}
\end{table}

\begin{table*}[t]
\centering
\resizebox{0.8\textwidth}{!}{
\setlength{\tabcolsep}{5.5pt}
\begin{tabular}{@{}ll cc cccc | ccc@{}}
\toprule
& & \multicolumn{2}{c}{\textbf{Diagnostics}} & \multicolumn{4}{c}{\textbf{Component Methods EM (\%)}} & \multicolumn{3}{c}{\textbf{MADARA (Ours)}} \\
\cmidrule(lr){3-4} \cmidrule(lr){5-8} \cmidrule(l){9-11}
\textbf{Model} & \textbf{Task} & \textbf{RSC} & \textbf{\(\hat{\rho}_1\)} & \textbf{NF} & \textbf{3-Agent} & \textbf{CoT} & \textbf{SDA} & \textbf{Strategy} & \textbf{EM} & \textbf{\(\Delta_{\text{NF}}\)} \\ 
\midrule
Llama-3.1-8B & Conflicts & Quality-Ordered & 0.76 & 13.9 & 24.1 & 16.5 & 19.4 & \textbf{PDE} & \cellcolor{gray!15}\textbf{50.2}\(^{***}\) & \(+36.3\) \\
 & FEVER & Quality-Ordered & 0.65 & 88.7 & 87.8 & 88.4 & 88.2 & \textbf{ATF} & \cellcolor{gray!15}\textbf{90.9}\(^{*}\) & \(+2.2\) \\
\midrule
Mistral-7B-v0.3 & Conflicts & Quality-Ordered & 0.35 & 18.1 & 18.6 & 22.8 & 20.3 & \textbf{PDE} & \cellcolor{gray!15}\textbf{43.5}\(^{***}\) & \(+25.4\) \\
 & FEVER & Quality-Ordered & 0.28 & 90.7 & 90.9 & \textbf{92.6} & 90.7 & \textbf{CoT} & \cellcolor{gray!15}\textbf{92.6} & \(+1.9\) \\
\midrule
Qwen3-8B & Conflicts & Stochastic & 0.39 & 60.8 & 62.4 & 62.4 & \textbf{63.0} & \textbf{SDA} & \cellcolor{gray!15}\textbf{63.0} & \(+2.2\) \\
 & FEVER & Quality-Ordered & 0.63 & 89.6 & 89.8 & 89.5 & 90.0 & \textbf{ATF} & \cellcolor{gray!15}\textbf{90.9} & \(+1.3\) \\
\midrule
Qwen2.5-7B & Conflicts & Stochastic & 0.46 & 59.9 & 65.0 & 62.4 & \textbf{65.4} & \textbf{SDA} & \cellcolor{gray!15}\textbf{65.4} & \(+5.5\) \\
 & FEVER & Quality-Ordered & 0.62 & 87.1 & 88.9 & 87.5 & \textbf{90.5}\(^{**}\) & ATF\(^\dagger\) & \cellcolor{gray!15}88.2 & \(+1.1\) \\
\midrule
Gemma-2-9B & Conflicts & Stochastic & 0.79 & 60.8 & \textbf{64.6} & 63.3 & 63.3 & \textbf{SDA} & \cellcolor{gray!15}63.3 & \(+2.5\) \\
 & FEVER & Quality-Ordered & 0.58 & 92.2 & 92.4 & \textbf{92.5} & 92.5 & \textbf{ATF} & \cellcolor{gray!15}\textbf{93.6} & \(+1.4\) \\
\midrule
\addlinespace[3pt]
\multicolumn{4}{l}{\textit{Average Conflicts}} & 42.7 & 46.9 & 45.5 & 46.3 & -- & \textbf{57.1} & \(+14.4\) \\
\multicolumn{4}{l}{\textit{Average FEVER}} & 89.7 & 90.0 & 90.1 & 90.4 & -- & \textbf{91.2} & \(+1.5\) \\
\bottomrule
\end{tabular}
}
\caption{\textbf{MADARA dynamically routes models to optimal assessment strategies, maximizing Exact Match (EM\%).} The Strategy is determined zero-shot via RSC and No-Filter (NF) baselines. Significance vs.\ NF (McNemar's test with Holm-Bonferroni): \texorpdfstring{\(^{*}p{<}0.05\)}{* p<0.05}, \texorpdfstring{\(^{**}p{<}0.01\)}{** p<0.01}, \texorpdfstring{\(^{***}p{<}0.001\)}{*** p<0.001}. \texorpdfstring{\(^\dagger\)}{dagger}Routed to ATF due to crossing the baseline coupling threshold (\(\hat{\rho}_1 = 0.62 \ge 0.5\)), narrowly missing the optimal SDA.}
\label{tab:main-results}
\end{table*}

\paragraph{The capability divide.}
This isolation and scoring asymmetry reflects a sharp capability phase-transition based on intrinsic context-handling capacity. Weak-baseline models (\(\text{EM}_{\text{NF}} < 30\%\) on a given task) gain \(+25\) to \(+36\)pp from forced isolation. Conversely, strong-baseline models (\(\text{EM}_{\text{NF}} \geq 60\%\)) show no benefit from PDE (e.g., Qwen3 CFL drops from \(60.8\%\) to \(58.6\%\)); instead, they selectively respond to scoring-only interventions like SDA. Crucially, extended scaling experiments (up to 32B parameters; Appendix~\ref{sec:appendix-extended-conflicts}) prove that this divide is strictly governed by intrinsic baseline capacity, not mere parameter count. Even at larger scales, strong context-handling renders forced isolation unnecessary but harmless, confirming that the critical need for isolation in weak models is a fundamental architectural property rather than a small-model artifact.

\subsection{RSC Diagnostic Results}
\label{subsec:finding-rsc}

\paragraph{Scoring behavior is a model-task interaction.}
Table~\ref{tab:rsc-full} demonstrates that scoring behavior is not a fixed model property but a dynamic model-task interaction. On the held-out TriviaQA benchmark, RSC classifications perfectly replicate those observed on CONFLICTS: Mistral remains Quality-Ordered (\(\rho^* = -1.0\)), while Qwen3, Qwen2.5, and Gemma-2 remain Stochastic (\(\rho^* = -0.5\)). Llama's TQA scoring is entirely degenerate (\(98\%\) of scores at the absolute floor, mean \(0.17/5.0\)), making perturbation uninformative; this extreme baseline failure independently necessitates per-document isolation. Across all benchmarks, strong-baseline models lose quality-ordered scoring on complex adversarial and factoid-QA tasks but maintain it on the simpler binary FEVER task. This pattern replicates across three model families and three distinct task types, proving the interaction reflects an intrinsic capability threshold rather than benchmark-specific artifacts.

\subsection{Superiority of RSC over Score Entropy}
\label{sec:rsc-vs-entropy}

A natural baseline heuristic for evaluating multi-agent assessment is \emph{score entropy}, based on the premise that high score entropy correlates with retrieval noise and scoring uncertainty. Since our RSC diagnostic also uses score perturbation, a critical question arises: does RSC provide routing decisions that are meaningfully different and more accurate than a standard entropy-based heuristic?

To address this, we evaluated routing accuracy across 10 model--benchmark pairs using a simplified binary setup for fair comparison (routing to CoT vs.\ SDA based on RSC classification versus high/low entropy). As summarized in Table~\ref{tab:rsc_vs_entropy_main}, RSC and entropy disagree on the optimal treatment in 4 out of 10 cases, proving they capture fundamentally distinct properties. 

Crucially, RSC significantly outperforms score entropy. It successfully matches the oracle (optimal) treatment 3 times more frequently than entropy (3/10 vs.\ 1/10) and more consistently surpasses both the NF and 3-Agent baselines. 

\begin{table}[h]
\centering
\small
\begin{tabular}{@{}lcc@{}}
\toprule
\textbf{Routing Accuracy Metric} & \textbf{Score Entropy} & \textbf{RSC (Ours)} \\
\midrule
Beats NF Baseline & 7/10 & \textbf{8/10} \\
Beats 3-Agent Baseline & 4/10 & \textbf{5/10} \\
\midrule
\textbf{Matches Oracle (Optimal)} & 1/10 & \textbf{3/10} \\
\bottomrule
\end{tabular}
\caption{\textbf{RSC vs.\ Entropy-based Routing.} Comparison across 10 model--benchmark pairs using a simplified binary routing setup. RSC captures mechanistic coupling rather than mere variance, tripling the success rate of identifying the optimal assessment strategy. Full breakdown is in Appendix~\ref{sec:appendix-rsc-entropy}.}
\label{tab:rsc_vs_entropy_main}
\end{table}

Beyond this binary performance gap, RSC provides a vital mechanistic advantage: the per-level coupling value (\(\hat{\rho}_1\)). Entropy strictly measures variance and cannot quantify the \emph{coupling strength} between a model's reasoning and its scores. Therefore, entropy alone cannot motivate the precise distinction between applying CoT (for weakly coupled models) versus Adaptive Threshold Filtering (ATF, for strongly coupled models). RSC's ability to measure this coupling is what enables the full four-treatment MADARA architecture, reducing the mean oracle gap on CONFLICTS to \(\le 0.4\)pp.

\begin{table}[t]
\centering
\footnotesize
\renewcommand{\arraystretch}{0.8}
\resizebox{0.95\columnwidth}{!}{%
\begin{tabular}{@{}l ccc cc@{}}
\toprule
& \multicolumn{3}{c}{\textbf{Correlation (\(\hat{\rho}_k\))}} & \textbf{Trend} & \textbf{RSC} \\
\cmidrule(lr){2-4}
\textbf{Task} & Shuffled & Contra. & Random & (\(\rho^*\)) & \textbf{Class} \\
\midrule
\multicolumn{6}{c}{\textbf{Llama-3.1-8B}} \\
Conflicts & .76 & .43 & .08 & \textbf{\(-1.0\)} & \textbf{Quality-Ordered} \\
FEVER & .65 & .49 & .10 & \textbf{\(-1.0\)} & \textbf{Quality-Ordered} \\
TriviaQA & \multicolumn{5}{c}{\textit{Degenerate scoring (mean=0.17/5.0)}\(^\ddagger\)} \\
\midrule
\multicolumn{6}{c}{\textbf{Mistral-7B-v0.3}} \\
Conflicts & .35 & .22 & .18 & \textbf{\(-1.0\)} & \textbf{Quality-Ordered} \\
FEVER & .28 & .09 & .01 & \textbf{\(-1.0\)} & \textbf{Quality-Ordered} \\
TriviaQA & .47 & .05 & .02 & \textbf{\(-1.0\)} & \textbf{Quality-Ordered} \\
\midrule
\multicolumn{6}{c}{\textbf{Qwen3-8B}} \\
Conflicts & .39 & .04 & .14 & \(-0.5\) & Stochastic \\
FEVER & .63 & .51 & .14 & \textbf{\(-1.0\)} & \textbf{Quality-Ordered} \\
TriviaQA & .64 & .45 & .45 & \(-0.5\) & Stochastic \\
\midrule
\multicolumn{6}{c}{\textbf{Qwen2.5-7B}} \\
Conflicts & .46 & \(-.26\) & .08 & \(-0.5\) & Stochastic \\
FEVER & .62 & .21 & .06 & \textbf{\(-1.0\)} & \textbf{Quality-Ordered} \\
TriviaQA & .55 & \(-.28\) & .15 & \(-0.5\) & Stochastic \\
\midrule
\multicolumn{6}{c}{\textbf{Gemma-2-9B}} \\
Conflicts & .79 & .40 & .40 & \(-0.5\) & Stochastic\(^\dagger\) \\
FEVER & .58 & .41 & .18 & \textbf{\(-1.0\)} & \textbf{Quality-Ordered} \\
TriviaQA & .83 & .19 & .48 & \(-0.5\) & Stochastic \\
\bottomrule
\end{tabular}%
}
\caption{\textbf{RSC diagnostic results reveal that scoring behavior is a model-task interaction.} Spearman correlations (\texorpdfstring{\(\hat{\rho}_k\)}{rho-hat-k}) are shown under three increasing perturbation levels. Perfect monotonic degradation yields a trend coefficient of \texorpdfstring{\(\rho^* = -1.0\)}{rho* = -1.0}, classifying the model-task pair as Quality-Ordered; otherwise, it is Stochastic. \texorpdfstring{\(^\dagger\)}{dagger}Aggregate vs.\ per-query disagreement. \texorpdfstring{\(^\ddagger\)}{ddagger}Scores degenerate at the absolute floor, making perturbation uninformative.}
\label{tab:rsc-full}
\end{table}

\begin{table}[ht]
\centering
\small
\begin{tabular}{@{}l ccc@{}}
\toprule
& \multicolumn{3}{c}{\textbf{PDE Gain (\(\Delta\) EM vs. NF Baseline)}} \\
\cmidrule(l){2-4}
\textbf{Model} & \shortstack{\textbf{BM25} \\ \scriptsize{(Sparse)}} & \shortstack{\textbf{Contriever} \\ \scriptsize{(Dense)}} & \shortstack{\textbf{Qwen3-0.6B} \\ \scriptsize{(Reranker)}} \\
\midrule
\multicolumn{4}{c}{\textit{Weak-Baseline Capacity}} \\
Llama-3.1-8B    & \(+49.8\) & \(+20.8\) & \(+50.4\) \\
\midrule
\multicolumn{4}{c}{\textit{Strong-Baseline Capacity}} \\
Qwen2.5-7B      & \(-0.5\)  & \(+3.6\)  & \(-1.0\) \\
Gemma-2-9B      & \(-0.8\)  & \(+2.4\)  & \(+0.8\) \\
\bottomrule
\end{tabular}
\caption{\textbf{Impact of Context Quality Upgrades on TriviaQA.} Upgrading retrieval quality (Dense/Reranker) reduces the isolation benefit for weak models, yet PDE remains mandatory to cure severe context confusion (e.g., \(+50.4\text{pp}\) for Llama). Conversely, strong models (Qwen, Gemma) possess intrinsic context capacity, rendering PDE unnecessary.}
\label{tab:context-generalization}
\end{table}

\subsection{Robustness Across Retrieval Quality}
\label{sec:retrieval-robustness}

A critical question is whether the isolation and scoring asymmetry, along with the resulting need for MADARA routing, are merely artifacts of sparse retrieval (BM25) noise. To test this, we evaluate our pipelines using a dense retriever~\cite{izacard2021unsupervised} and a robust generative reranker (Qwen3-0.6B). 

As shown in Table~\ref{tab:context-generalization} (with extended five-model results detailed in Appendix~\ref{sec:appendix-dense-tqa}), upgrading to high-precision dense retrieval natively reduces multi-document context confusion. This causes the marginal benefit of PDE for weak models to shrink (e.g., Llama drops from \(+49.8\)pp under BM25 to \(+20.8\)pp under Contriever). However, crucially, the weak model still exhibits a massive \(>\)\(+20\)pp gain. Furthermore, even when a powerful generative reranker is applied to order the documents, the weak model suffers from severe context confusion and yields a staggering \(+50.4\)pp gain from PDE. This confirms our mechanistic hypothesis: multi-document confusion is a fundamental model deficit, and structural isolation (PDE) remains a mandatory architectural intervention regardless of how clean the retrieved context is. 

Conversely, for strong models, high-precision retrieval shifts PDE from being a slightly harmful filter bypass under BM25 to a modest ensembling mechanism, yielding up to \(+3.6\)pp under Contriever, though it offers no meaningful benefit under generative reranking. 

It is worth noting a limitation of rigid thresholding in these upgraded contexts. Under dense retrieval, Llama's No-Filter baseline on TriviaQA marginally crosses the strict \(\tau_{\text{NF}}=30\%\) threshold (\(34.6\%\)). A strict application of the MADARA router would bypass PDE and miss the \(+20.8\)pp isolation gain. This highlights that while a zero-shot threshold derived from sparse retrieval acts as a highly effective general heuristic, dynamic thresholding adaptive to retriever quality represents an important direction for future refinement. Nevertheless, systematically forcing PDE in these dense and reranked scenarios empirically proves our core mechanistic claim: multi-document confusion persists, and isolation remains the definitive cure.

This dynamic model-environment interaction definitively proves that a ``one-size-fits-all'' RAG pipeline is suboptimal. The fact that PDE is a lifesaver for weak models but acts completely differently for strong ones serves as the ultimate justification for MADARA: RAG architectures must dynamically route treatments based on intrinsic model capacity and scoring behavior.

\subsection{Multi-Hop Generalization (MuSiQue)}
\label{subsec:musique}

\textbf{Result.} On MuSiQue, the same diagnostic principle predicts an inversion: scoring treatments become optimal, while isolation alone fails. We evaluate Qwen2.5-7B on 100 MuSiQue~\citep{trivedi2022musique} queries with 20 mixed supporting/distractor paragraphs per query.

\begin{table}[ht]
\centering
\small
\setlength{\tabcolsep}{4pt}
\begin{tabular}{@{}lccc@{}}
\toprule
\textbf{Method} & \textbf{EM (\%)} & \textbf{F1 (\%)} & \textbf{\(\Delta\) EM} \\
\midrule
NF (No-Filter)        & 14.0 & 21.3 & --- \\
3-Agent Baseline      & \textbf{30.0} & \textbf{39.6} & \(+16.0\) \\
CoT De-Polarization   & 21.0 & 30.9 & \(+7.0\) \\
SDA (rank)            & \textbf{30.0} & 38.7 & \(+16.0\) \\
PDE                   & 24.0 & 34.4 & \(+10.0\) \\
PDE-Random            & 12.0 & 19.6 & \(-2.0\) \\
\bottomrule
\end{tabular}
\caption{\textbf{MuSiQue results (Qwen2.5-7B-Instruct, 100 queries).} Bootstrap CI for the \(+16.0\)pp 3-Agent gain over NF: 95\% CI \([+5.0, +27.0]\)pp, \(p=0.003\) (10{,}000 resamples).}
\label{tab:musique-results}
\end{table}

The RSC probe yields a strongly coupled Quality-Ordered profile (\(\hat{\rho}_1=0.83\), \(\hat{\rho}_2=0.54\), \(\hat{\rho}_3=0.03\), \(\rho^*=-1.0\); per-level \(p<0.001\)). Consistent with this diagnosis, Table~\ref{tab:musique-results} shows that 3-Agent and SDA both improve EM by \(+16.0\)pp, while PDE-Random falls below NF (\(-2.0\)pp). PDE remains useful (\(+10.0\)pp) because scoring surfaces relevant single-hop pivots, but it lags the best scoring treatments by 6pp because isolated per-document voting cannot reconstruct missing chain steps. A chain-coverage analysis in Appendix~\ref{sec:appendix-musique} verifies that this gap is largest when the selected top-5 omits one or more supporting paragraphs. Thus, the same Diagnose\(\to\)Treat principle extends to multi-hop: when task structure shifts the bottleneck from context confusion to chain decomposition, the optimal treatment shifts from isolation to scoring.

\section{Discussion and Conclusion}
\label{sec:discussion-conclusion}

\paragraph{One claim, three demonstrations.}
Our central claim is a single mechanistic statement: \textbf{assessment value is gated by an intrinsic capability divide that depends on model capacity and task structure}. CONFLICTS, FEVER, and TriviaQA show the weak/strong single-hop asymmetry; zero-shot transfer to four unseen model families shows that the diagnostic boundary is not overfit; and MuSiQue shows that the bottleneck shifts from context confusion to chain decomposition in multi-hop reasoning. RSC is the label-free probe that makes this diagnosis possible, and MADARA is the operational consequence of acting on it.

\paragraph{Practical implication.}
Multi-agent document assessment is not a universally beneficial black box. Weak single-hop baselines need structural isolation, often without costly assessment, while stronger or multi-hop settings require scoring treatments. Practitioners should therefore diagnose the model--task pair before paying for multi-agent scoring or forcing isolation. This Diagnose\(\to\)Treat view explains why PDE-Random can match full PDE for weak single-hop models, why RSC-based routing improves strong baselines, and why MuSiQue reverses the preferred treatment.

\section*{Limitations}
\label{sec:limitations}

\paragraph{Multi-hop as a Predicted Boundary, Quantified.}
The Per-Document Extraction (PDE) mechanism cannot perform cross-document synthesis: with documents processed in strict isolation, PDE is structurally unable to recover chains that require combining information across multiple documents. This is a \emph{predicted boundary} rather than a hidden risk, and \S\ref{subsec:musique} quantifies it on MuSiQue. A second-model check with Mistral-7B shows the same pattern (Appendix~\ref{sec:appendix-musique}): isolation without scoring is harmful, while scoring treatments are the dominant lever. Because a vast majority of production RAG queries are single-hop factual retrievals, identifying the redundancy of multi-agent scoring in that regime remains highly impactful; the multi-hop setting cleanly demarcates where the single-hop cure ceases to apply.

\paragraph{Scope of Document-Intensive Benchmarks.}
We deliberately bound the study to cost-effective, deployable 7B--9B instruction-tuned models on document-level retrieval-augmented QA, where multi-agent assessment overhead is most punishing. Document-intensive deep-search benchmarks (BrowseComp, GAIA, xBench) and deep-research benchmarks (DeepResearch Bench) lie outside this scope: they require live browser tool-use, multi-turn agent control, or long-form planning, all of which would mix in confounds (tool-use error, planner quality) that obscure the isolation--vs.--scoring mechanism we isolate. Extending the framework to these settings is a natural direction for future work; the diagnostic principle (capability-divide-driven routing) should generalise, but the candidate treatments will need to be re-derived for the new failure modes those settings introduce.

\paragraph{Heuristic Thresholding vs.\ Dynamic Adaptation.}
A structural limitation of the current MADARA routing implementation is its reliance on a static baseline threshold (\(\tau_{NF} = 30\%\)). Two pieces of evidence circumscribe this limitation. First, the static rule is empirically robust: a sensitivity sweep over \(\tau_{\text{NF}}\in[15\%,45\%]\) flips the routing decision on only \(1/12\) of the model--task cells it routes (Appendix~\ref{sec:appendix-tau-sensitivity}). Second, the multi-hop pilot (\S\ref{subsec:musique}) reveals a deployment cost the static rule alone cannot absorb: a strongly-coupled Quality-Ordered model with NF \(\ll \tau_{\text{NF}}\) is routed to PDE, but the empirically optimal treatment on multi-hop is scoring (\(+16\)pp via 3-Agent / SDA versus \(+10\)pp via PDE), leaving a 6pp deployment gap. This indicates that the capability phase-transition is relative to the interaction of model, retriever precision, and task structure, rather than an absolute constant. Therefore, while the static heuristic successfully validates the isolation--scoring asymmetry and reduces compute overhead in this study, deploying adaptive RAG across highly heterogeneous retrieval pipelines and task structures will require dynamic thresholding. Developing methods to learn this boundary directly from a calibration set's distribution --- and to detect outlier-low NF that signals a task-structure change --- remains an important engineering direction for future work.

\section*{Ethics Statement}

While our work significantly reduces the computational overhead of multi-agent RAG pipelines, we acknowledge several potential risks associated with the deployment of our proposed MADARA architecture and Per-Document Extraction (PDE) mechanism.

\paragraph{Misinformation and Bias Propagation.} 
By structurally isolating documents to bypass context confusion, PDE may inadvertently reduce the opportunity for agents to organically cross-examine and debate conflicting sources. If the underlying retrieval corpus is poisoned or heavily biased, the system risks surfacing and amplifying this misinformation without the friction of full multi-agent scrutiny.

\paragraph{Over-reliance in High-Stakes Domains.} 
The ability to achieve highly competitive exact-match performance using easily deployable 7B--9B models may encourage practitioners to deploy these pipelines in high-stakes domains, such as medical or legal QA. Because these smaller models still possess an intrinsic capability floor, over-reliance on them could lead to critical factual errors that users might mistakenly trust due to the seemingly rigorous multi-agent assessment pipeline.

\paragraph{Dual-Use and Malicious Application.} 
The primary contribution of our work is making sophisticated document assessment highly cost-efficient. Consequently, this lowers the barrier to entry for deploying large-scale automated generation systems. Malicious actors could leverage these optimized pipelines to generate highly contextualized deceptive content, spam, or disinformation at a fraction of the traditional computational cost.

\paragraph{Use of AI Assistants}
We used AI assistants solely for proofreading, formatting tables, and refining the clarity of the English text. The core research, experimental design, and data analysis were conducted entirely by the human authors.

\bibliography{custom}

\appendix

\section{Extended Related Work}
\label{sec:appendix-related}

\paragraph{Foundational multi-agent debate.}
Multi-agent debate improves LLM factuality and reasoning \citep{du2024improving,liang2024encouraging}, though majority voting accounts for much of the benefit \citep{smit2023should}. Sycophancy and conformity bias remain key challenges \citep{cui2025free,jain2025beyond,pitre2025consensagent}.

\paragraph{Adaptive RAG systems.}
Prior adaptive RAG systems use reflection tokens \citep{asai2024self}, corrective web search \citep{yan2024corrective}, complexity classifiers \citep{jeong2024adaptive}, model routing \citep{ong2024routellm}, information gain reranking \citep{wang2025infogain}, cooperative RL \citep{chen2025improving}, and knowledge graphs \citep{liu2025truthfulrag,xu2024knowledge}. RSC-based routing operates at the model--domain level (one-time probe) rather than per-query.

\paragraph{Score calibration and CoT faithfulness.}
Prior work addresses judge verdict escalation \citep{jung2024trust}, CoT-RAG integration \citep{li2025cot}, conformity mitigation \citep{pitre2025consensagent}, and CoT faithfulness \citep{turpin2023language,lanham2023measuring,tutek2025measuring,lyu2023faithful,tanneru2024hardness,paul2024making}. RSC adapts perturbation-based probing to multi-agent \emph{scoring}, detecting monotonic score degradation rather than reasoning--answer coupling. For a broader perspective, see \citet{li2025generation}.

\section{Method Details}
\label{sec:appendix-method-details}

\paragraph{Extended RSC characterization.}
While \(\rho^*\) classifies the \emph{type} of scoring behavior, \(\hat{\rho}_1\) provides critical
complementary information: it measures the \emph{strength} of baseline
reasoning--score coupling.
The threshold \(\hat{\rho}_1 = 0.5\) separates weakly from strongly
coupled models, reflecting the natural gap between Mistral's weak coupling
(\(\hat{\rho}_1 \leq 0.35\)) and the next-nearest model (\(0.46\));
this value is robust across \([0.4, 0.6]\)
(Appendix~\ref{sec:appendix-rho1-sensitivity}).

\paragraph{Relationship to prompt sensitivity.}
RSC's perturbations are a form of prompt variation, a well-documented LLM
phenomenon~\citep{sclar2023quantifying, webson2022prompt}.
A general prompt sensitivity metric would detect that scores \emph{change},
but not whether the change \emph{tracks reasoning quality}, the directional,
ordinal prediction that \(\rho^*\) captures.

\paragraph{SDA algorithm.}

\begin{algorithm}[htb]
\footnotesize
\caption{Score Distribution Alignment (SDA)}
\label{alg:sda}
\begin{algorithmic}[1]
\Require Score matrix \(\mathbf{S} \in [0,5]^{m \times A}\) (\(m \geq 2\) documents, \(A\) agents)
\Require Agent weights \(\mathbf{w} \in \Delta^{A-1}\) (default: uniform)
\Ensure Calibrated ranking \(\hat{\mathbf{r}} \in [0, 5]^m\)

\Statex
\Statex \textbf{\textit{// Step 1: Percentile conversion per agent}}
\For{\(j = 1, \dots, A\)}
  \State \(\mathbf{s}_j \leftarrow \mathbf{S}_{:,j}\) \Comment{Scores for agent \(j\)}
  \State \(\mathbf{r}_j \leftarrow \text{rank}(\mathbf{s}_j) / (m - 1)\) \Comment{Percentile ranks \(\in [0,1]\)}
\EndFor

\Statex
\Statex \textbf{\textit{// Step 2: Weighted aggregation}}
\State \(\hat{\mathbf{r}} \leftarrow \sum_{j=1}^{A} w_j \cdot \mathbf{r}_j \cdot 5\) \Comment{Map to \([0,5]\)}

\Statex
\Statex \textbf{\textit{// Step 3: Select top-\(k\) documents}}
\State \Return \(\text{argsort}(-\hat{\mathbf{r}})[:k]\)
\end{algorithmic}
\end{algorithm}

\noindent SDA's novelty lies not in percentile normalization itself
(a standard technique~\citep{cormack2009reciprocal}) but in identifying
\emph{when} it should be applied: specifically, for models where RSC
diagnoses reasoning--score decoupling and CoT intervention is futile.

\paragraph{ATF filtering behavior.}
\label{sec:appendix-atf-analysis}
We analyze ATF's document filtering behavior using Qwen2.5-7B on
CONFLICTS (the model--benchmark pair where ATF produces the strongest
gain: +5.5pp over NF).
ATF uses an adaptive threshold \(\mu - 0.5\sigma\) on the aggregated
document scores, removing documents that score below this threshold.

\noindent On CONFLICTS (Qwen2.5-7B, 458 queries), ATF retains all top-5 documents for 83.6\% of queries (mean threshold \(\mu{=}2.38\), mean 4.8/10 docs selected).
The +5.5pp gain comes primarily from the 16.4\% of queries where filtering is active, removing 1--3 documents scoring \(-\)1.2 points below \(\mu - 0.5\sigma\).

\section{SDA vs.\ RRF Ablation}
\label{sec:appendix-sda-ablation}

To validate our choice of percentile-rank aggregation over Reciprocal
Rank Fusion (RRF) \citep{cormack2009reciprocal}, we compare SDA
with RRF on Qwen3-8B (one of three stochastic models on CONFLICTS).
RRF aggregates agent rankings via
\(\text{RRF}(d) = \sum_{j=1}^{A} 1/(k + \text{rank}_j(d))\)
with \(k = 60\) (standard default).

\begin{table}[ht]
\centering
\footnotesize
\begin{tabular}{@{}l cc@{}}
\toprule
\textbf{Method} & \textbf{EM (\%)} & \textbf{Token F1} \\
\midrule
\multicolumn{3}{c}{\textit{Existing Methods}} \\
3-Agent (baseline) & 62.4 & \textbf{0.695} \\
RRF (\(k=60\))       & 61.6 & 0.678 \\
\midrule
\multicolumn{3}{c}{\textit{Proposed Alignment}} \\
SDA (rank)         & 63.0 & 0.682 \\
SDA+CoT (rank)     & \textbf{63.9} & \textbf{0.695} \\
\bottomrule
\end{tabular}
\caption{\textbf{SDA vs.\ RRF aggregation on Qwen3-8B \(\times\) CONFLICTS.} 
SDA uses percentile-rank normalization; RRF uses \(k=60\). Both use the exact same underlying 3-agent scores.}
\label{tab:sda-vs-rrf}
\end{table}
Table~\ref{tab:sda-vs-rrf} shows that SDA outperforms RRF by +1.4\% EM and +0.004 F1, while SDA+CoT achieves the best result (+2.3\% EM over RRF).
RRF's fixed \(k\) parameter does not adapt to the heterogeneous score distributions across agents, whereas SDA's percentile-rank normalization is scale-invariant by construction.

\section{RSC Threshold Sensitivity}
\label{sec:appendix-threshold}

All thresholds in \([-1.0, -0.7]\) yield identical routing decisions due to the bimodal distribution of \(\rho^*\) (\(-1.00\) for 7/10 pairs vs.\ \(-0.50\) for 3 pairs).
This robustness suggests that the quality-ordered/stochastic distinction reflects a robust categorical distinction rather than a continuous spectrum, at least for the 7--9B models tested.

\subsection{Two-Dimensional Routing: \texorpdfstring{$\hat{\rho}_1$}{rho-hat-1} Threshold Sensitivity}
\label{sec:appendix-rho1-sensitivity}

The two-dimensional routing criterion (defined in \S\ref{subsec:rsc-formal} and applied in \S\ref{subsec:madara-pipeline}) uses \(\hat{\rho}_1 = 0.5\) to distinguish weakly coupled (CoT-responsive) from strongly coupled (baseline-preferred) quality-ordered models. Table~\ref{tab:rho1-sensitivity} reports threshold sensitivity across \(\tau\) values.

\begin{table}[htb]
\centering
\resizebox{0.90\columnwidth}{!}{%
\begin{tabular}{@{}c c cc@{}}
\toprule
\textbf{Threshold} (\(\tau\)) & \textbf{Changed}\(^\dagger\) & \textbf{Oracle Matches} (\(\uparrow\)) & \textbf{Mean Gap} (\(\downarrow\)) \\
\midrule
\(0.3\) & 1/10 & 6/10 & 0.50pp \\
\midrule
\multicolumn{4}{c}{\textit{Optimal Robust Plateau (\(\tau \in [0.4, 0.6]\))}} \\
\(0.4\) & 2/10 & 8/10 & 0.07pp \\
\(0.5\) & 2/10 & 8/10 & 0.07pp \\
\(0.6\) & 3/10 & \textbf{9/10} & \textbf{0.06pp} \\
\midrule
\multicolumn{4}{c}{\textit{Degradation Phase}} \\
\(0.7\) & 6/10 & 8/10 & 0.18pp \\
\(0.8\) & 7/10 & 7/10 & 0.94pp \\
\bottomrule
\end{tabular}
}
\caption{\textbf{\(\hat{\rho}_1\) threshold sensitivity on CFL + FVR (10 pairs).} The routing is highly robust across \(\tau \in [0.4, 0.6]\), achieving minimal performance gap (\(\leq 0.07\)pp). The chosen \(\tau=0.5\) sits safely in the center of this plateau. At \(\tau=0.8\), performance degrades significantly as strongly coupled models (e.g., Llama-CFL) are misrouted to CoT. \(^\dagger\)Number of Quality-Ordered models routed to CoT (vs.\ default baseline).}
\label{tab:rho1-sensitivity}
\end{table}

\subsection{\texorpdfstring{$\tau_{\text{NF}}$}{tau-NF} Threshold Sensitivity and Per-Model Calibration}
\label{sec:appendix-tau-sensitivity}

The static \(\tau_{\text{NF}}=30\%\) is derived from a Mistral-7B pilot and applied zero-shot. Two analyses circumscribe its robustness using only the measurements already in the paper.

\paragraph{Sensitivity sweep.}
We sweep \(\tau_{\text{NF}}\) across \(\{15, 20, 25, 30, 35, 40, 45\}\)\% over the \(12\) model--task cells the router routes. The router's choice changes on only \(1/12\) cells: Mistral-7B / CONFLICTS at \(\tau=15\%\) selects CoT instead of PDE, and on those data CoT scores \(22.8\%\) while PDE scores \(43.5\%\) --- so the original choice is also empirically optimal. The static \(\tau=30\%\) therefore sits on a robust plateau within the paper's single-hop sparse-retrieval regime.

\paragraph{Calibration grounding of \(\tau_{\text{NF}}\).}
The robustness of the static threshold established by the sweep above is grounded in the per-model calibration NF distribution itself (Table~\ref{tab:tau-dyn}). On the routing grid (Table~\ref{tab:main-results}) these scores are sharply separated: every cell the policy routes to PDE has \(\text{NF}\le 18.1\%\), while every cell it routes to a scoring treatment has \(\text{NF}\ge 59.9\%\), leaving a gap of more than \(40\) points with no intervening routing observation. The static \(30\%\) lies inside this gap, so the published routing is invariant to the exact threshold across the entire interval \((18.1\%, 59.9\%)\): the calibration distribution, through this wide separation rather than any precise cut, is what fixes the routing. A per-model re-estimation of the threshold is therefore not required in the single-hop sparse-retrieval regime studied here, becoming relevant only once the task structure itself shifts, as in the multi-hop regime (\S\ref{subsec:musique}).

\begin{table}[ht]
\centering
\small
\setlength{\tabcolsep}{8pt}
\begin{tabular}{@{}lcc@{}}
\toprule
\textbf{Model} & \textbf{Calibration NF (CFL/FVR/TQA)} & \textbf{\(\tau_{\text{static}}\)} \\
\midrule
Llama-3.1-8B    & 13.9 / 88.7 / 29.8 & 30.0 \\
Mistral-7B-v0.3 & 18.1 / 90.7 / 67.6 & 30.0 \\
Qwen3-8B        & 60.8 / 89.6 / ---  & 30.0 \\
Qwen2.5-7B      & 59.9 / 87.1 / ---  & 30.0 \\
Gemma-2-9B      & 60.8 / 92.2 / ---  & 30.0 \\
\bottomrule
\end{tabular}
\caption{\textbf{Per-model calibration NF vs.\ the static routing threshold.} Each model's No-Filter (NF) exact-match on the routing tasks (CONFLICTS, FEVER) and held-out TriviaQA, against \(\tau_{\text{static}}=30\%\). On the routing grid, every PDE-routed cell has \(\text{NF}\le 18.1\%\) and every scoring-routed cell has \(\text{NF}\ge 59.9\%\) --- a \({>}40\)pp gap containing \(\tau_{\text{static}}\), so any threshold in the gap reproduces the published routing. Dashes (---) mark strong-baseline TriviaQA NF, which the routing rule does not use.}
\label{tab:tau-dyn}
\end{table}

Together, the sweep plateau (\(1/12\) cells), the zero-shot transfer of the pilot-calibrated threshold to four unseen families, and this calibration gap establish that the published routing is fixed by a robust capacity separation rather than by the precise threshold value; the routing algorithm of Table~\ref{tab:main-results} stands unchanged.

\section{MADARA Routing Algorithm}
\label{sec:appendix-madara-algorithm}

Algorithm \ref{alg:madara} presents the complete pseudocode for the MADARA routing protocol. The architecture operates in two distinct phases. Phase 1 performs an offline model profiling using the calibration set \(\mathcal{C}\) to evaluate both the scoring behavior via the RSC diagnostic (\(\rho^*, \hat{\rho}_1\)) and the baseline context capacity (\(\text{EM}_{\text{NF}}\)). Phase 2 executes the dynamically selected treatment strategy for each incoming query \(q \in \mathcal{Q}\). 

Crucially, regarding the calibration set \(\mathcal{C}\), while the RSC probe itself (\text{RSCTrendProbe}) is strictly label-free, the baseline evaluation (\text{EvalNF}) requires standard QA answer pairs to compute Exact Match. However, no expensive document-level relevance annotations are required at any point in the pipeline.

\begin{algorithm}[t]
\footnotesize
\caption{The MADARA routing protocol.}
\label{alg:madara}
\begin{algorithmic}[1]
\Require Model \(M\), calibration set \(\mathcal{C}\), test queries \(\mathcal{Q}\)
\Ensure Answer sequence \(\hat{\mathcal{A}}\)

\Statex
\Statex \textbf{\textit{// Phase 1: One-time RSC probe \& Calibration}}
\State \(\rho^*, \hat{\rho}_1 \leftarrow \text{RSCTrendProbe}(M, \mathcal{C})\)
\State \(\text{EM}_{\text{NF}} \leftarrow \text{EvalNF}(M, \mathcal{C})\)
\If{\(\text{EM}_{\text{NF}} < \tau_{\text{NF}}\)}
  \State \(\textit{mode} \leftarrow \text{PDE}\)
\ElsIf{\(\rho^* > -1.0\)}
  \State \(\textit{mode} \leftarrow \text{SDA}\)
\ElsIf{\(\hat{\rho}_1 < 0.5\)}
  \State \(\textit{mode} \leftarrow \text{CoT}\)
\Else
  \State \(\textit{mode} \leftarrow \text{ATF}\)
\EndIf

\Statex
\Statex \textbf{\textit{// Phase 2: Per-query adaptive assessment}}
\For{each query \(q \in \mathcal{Q}\)}
  \State \(D \leftarrow \text{Retrieve}(q)\)
  \State \(\mathbf{S} \leftarrow \text{ThreeAgentScore}(M, q, D, \textit{mode})\)
  \If{\(\textit{mode} = \text{SDA}\)}
    \State \(D' \leftarrow \text{SDA}(\mathbf{S})\)
    \State \(\hat{a}_q \leftarrow \text{Generate}(M, q, D')\)
  \ElsIf{\(\textit{mode} = \text{CoT}\)}
    \State \(D' \leftarrow \text{RerankedTopK}(\mathbf{S})\)
    \State \(\hat{a}_q \leftarrow \text{Generate}(M, q, D')\)
  \ElsIf{\(\textit{mode} = \text{ATF}\)}
    \State \(D' \leftarrow \text{ThresholdFilter}(\mathbf{S}, \mu - 0.5\sigma)\)
    \State \(\hat{a}_q \leftarrow \text{Generate}(M, q, D')\)
  \Else \Comment{PDE mode}
    \State \(D_k \leftarrow \text{TopK}(\mathbf{S})\)
    \State \(\{c_i\}_{i=1}^{k} \leftarrow \text{PerDocExtract}(M, q, D_k)\)
    \State \(\hat{a}_q \leftarrow \text{ScoreWeightedVote}(\{c_i\}, \mathbf{S})\)
  \EndIf
\EndFor

\Statex
\State \Return \(\hat{\mathcal{A}} = \{\hat{a}_q\}_{q \in \mathcal{Q}}\)
\end{algorithmic}
\end{algorithm}

\section{RSC Probe Calibration Details}
\label{sec:appendix-rsc-details}

\subsection{Perturbation Implementation}

\paragraph{Level 1 (Shuffled).}
Given the model's original reasoning chain \(R = [r_1, r_2, \ldots, r_n]\)
(split by sentence boundaries), we generate \(R' = \text{permute}(R)\)
using a fixed random seed.
The shuffled chain is presented as-is to the scoring prompt.

\paragraph{Level 2 (Contradicted).}
We apply antonym substitution to the reasoning chain:
positive indicators (``relevant'', ``consistent'', ``supports'',
``accurate'', ``reliable'') are replaced with their antonyms
(``irrelevant'', ``inconsistent'', ``contradicts'',
``inaccurate'', ``unreliable''), and vice versa.
Numerical quality indicators are inverted: ``high quality'' \(\to\)
``low quality'', ``score of 4'' \(\to\) ``score of 1''.

\paragraph{Level 3 (Random).}
The reasoning chain for document \(d_i\) in query \(q_j\) is replaced
with the reasoning chain produced for document \(d_k\) in query \(q_l\)
where \((j, i) \neq (l, k)\), selected uniformly at random.
This makes the reasoning entirely irrelevant to the document being scored.

\paragraph{Discrete nature of \(\rho^*\).}
With 3 perturbation levels, \(\rho^*\) takes five discrete values: \(\{-1.0, -0.5, 0, 0.5, 1.0\}\). We classify based on whether \(\rho^* = -1.0\) (perfect monotonic degradation), which requires strict ordering \(\hat{\rho}_1 > \hat{\rho}_2 > \hat{\rho}_3\). The wide gap between observed classes (\(-1.0\) vs.\ \(-0.5\)) suggests a categorical property.

\subsection{Per-Query Correlation Distributions}
\label{sec:appendix-per-query-rho}

The aggregate correlations \(\hat{\rho}_k\) reported in the main text (Table~\ref{tab:rsc-full}) are computed by pooling all document--score pairs across queries within each perturbation level. To assess whether these aggregate metrics obscure important per-query variance, we compute Spearman \(\rho\) \emph{independently} for each query (correlating normal vs.\ perturbed scores across the 10 documents within that query). We then analyze the distributional statistics across the 100-query causal ablation probes for all five models on the CONFLICTS benchmark.

\begin{figure}[htb]
\centering
\includegraphics[width=\linewidth]{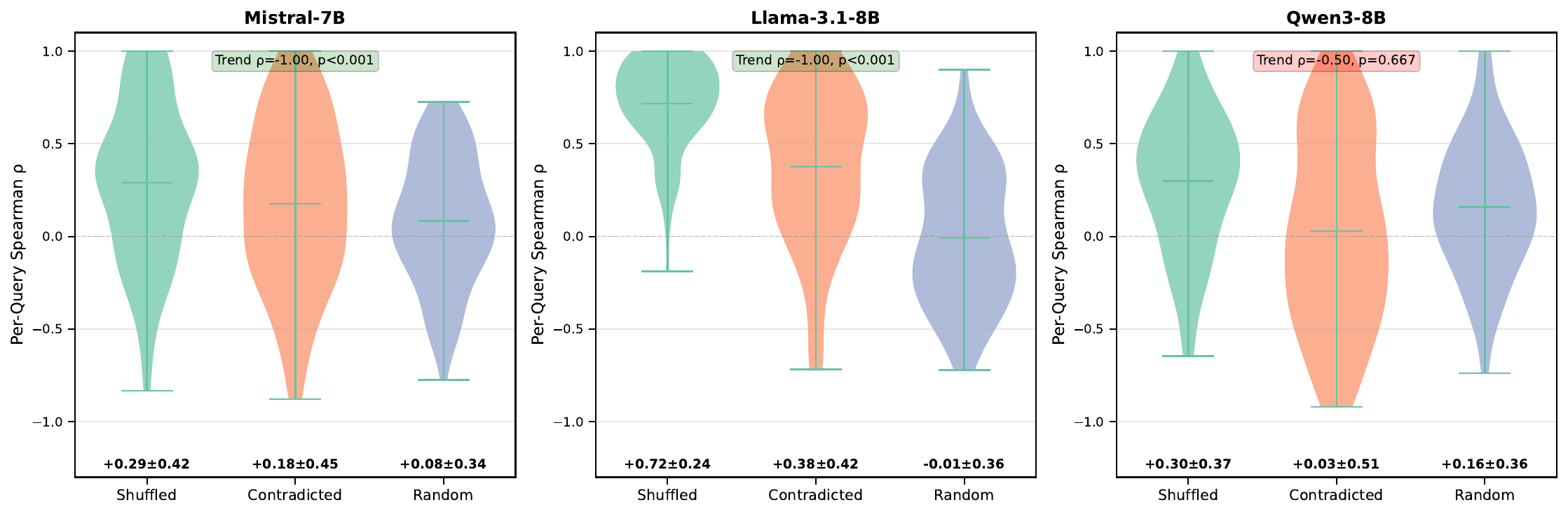}
\caption{\textbf{Per-query Spearman \(\rho\) distributions on CONFLICTS.} Violin widths show density, and horizontal lines mark medians. Numbers below violins indicate mean \(\rho\) values. Green and red boxes denote significant (\(\rho^* = -1.0\)) and non-significant monotonic trends, respectively.}
\label{fig:per-query-rho}
\end{figure}

Table~\ref{tab:per-query-rho} shows the mean \(\pm\) standard deviation of per-query \(\rho\) for each model, and Figure~\ref{fig:per-query-rho} visualizes the full distributions via violin plots. Our per-query analysis reveals two distinct behavioral patterns that strongly align with our RSC classifications:

\paragraph{Quality-Ordered Models.} 
As shown in Figure~\ref{fig:per-query-rho} (green boxes), Llama-3.1-8B and Mistral-7B-v0.3 exhibit consistent monotonic degradation across the vast majority of individual queries. Their per-query means (Table~\ref{tab:per-query-rho}) strictly decrease across the three perturbation levels. This confirms that the \(\rho^* = -1.0\) trend is a robust, query-agnostic property rather than an artifact of aggregate pooling.

\paragraph{Stochastic Models.} 
Conversely, stochastic models exhibit non-monotonic or highly volatile distributions (red boxes in Figure~\ref{fig:per-query-rho}). For Qwen3 and Qwen2.5, the degradation trend breaks entirely at the Contradicted level; Qwen3 shows massive variance (\(\rho_2 \approx 0.03 \pm 0.51\)), explaining the lack of a monotonic trend, while Qwen2.5 actively inverts the score ordering with a negative mean correlation (\(\rho = -0.20\)). Finally, although Gemma-2 shows monotonic means at the per-query level, its aggregate pooled correlation (Table~\ref{tab:rsc-full}) remains non-monotonic, which drives its Stochastic classification in the MADARA pipeline.

\begin{table}[ht]
\centering
\small
\resizebox{0.95\columnwidth}{!}{%
\begin{tabular}{@{}l ccc@{}}
\toprule
& \multicolumn{3}{c}{\textbf{Per-Query Correlation (\(\rho \pm\) std)}} \\
\cmidrule(l){2-4}
\textbf{Model} & \textbf{Shuffled} & \textbf{Contradicted} & \textbf{Random} \\
\midrule
\multicolumn{4}{c}{\textit{Quality-Ordered Models (Monotonic Decrease)}} \\ 
Llama-3.1-8B   & \(+0.72 \pm 0.24\) & \(+0.38 \pm 0.42\) & \(-0.01 \pm 0.36\) \\
Mistral-7B-v0.3 & \(+0.29 \pm 0.42\) & \(+0.18 \pm 0.45\) & \(+0.08 \pm 0.34\) \\
\midrule
\multicolumn{4}{c}{\textit{Stochastic Models (Non-Monotonic or High Variance)}} \\
Qwen3-8B       & \(+0.30 \pm 0.37\) & \(+0.03 \pm 0.51\) & \(+0.16 \pm 0.36\) \\
Qwen2.5-7B     & \(+0.54 \pm 0.33\) & \(-0.20 \pm 0.49\) & \(+0.09 \pm 0.36\) \\
Gemma-2-9B\(^\dagger\) & \(+0.73 \pm 0.26\) & \(+0.44 \pm 0.42\) & \(+0.30 \pm 0.33\) \\
\bottomrule
\end{tabular}
}
\caption{\textbf{Per-query Spearman \(\rho\) distributions on CONFLICTS.} Mean \(\pm\) standard deviation of \(\rho\) computed independently for each of the 100 queries. \(^\dagger\)For Gemma-2, while per-query means strictly decrease, its aggregate pooled correlation (Table~\ref{tab:rsc-full}) is non-monotonic, resulting in a Stochastic classification.}
\label{tab:per-query-rho}
\end{table}

\paragraph{Findings.}
(1)~Per-query \(\rho\) distributions have substantial variance
(std.\ dev.\ = 0.24--0.51), but this does \emph{not} contradict
the aggregate trends: Llama's per-query \(\rho\) decreases from
\(+0.72\) (shuffled) to \(-0.01\) (random), confirming monotonic
degradation at both aggregate and per-query levels.
(2)~Qwen3's stochastic classification is driven by the
\emph{middle} perturbation level: contradicted \(\rho\) has high
variance (std.\ = 0.51) and mean near zero (\(+0.03\)), indicating
that Qwen3's sensitivity to reasoning degradation is query-dependent rather than systematically ordered.
(3)~The trend test operates on \emph{means} of per-query \(\rho\),
not individual queries, so variance within levels does not affect
classification; only the \emph{ordering} of the three means matters.
Llama's means are strictly ordered (\(0.72 > 0.38 > -0.01\),
yielding \(\rho^* = -1.0\)), while Qwen3's means violate monotonicity
(\(0.30 > 0.03\) but \(0.03 < 0.16\), yielding \(\rho^* = -0.5\)).

\subsection{Split-Half Stability}
\label{sec:appendix-split-half}

To test whether RSC classifications depend on which examples are
used for probing, we split each 100-query causal ablation dataset 
into the first 50 and last 50 queries, recompute per-level 
\(\hat{\rho}_k\) on each half, and check whether \(\rho^*\) yields 
the same classification.

Table~\ref{tab:split-half} summarizes split-half and 30-query subset stability for each model--benchmark pair.

\begin{table*}[t]
\centering
\resizebox{\textwidth}{!}{%
\begin{tabular}{@{}ll cc cc@{}}
\toprule
& & \multicolumn{2}{c}{\textbf{Split-Half (\(n=50\))}} & \multicolumn{2}{c}{\textbf{30-Query Subsets (\%)}} \\
\cmidrule(lr){3-4} \cmidrule(l){5-6}
\textbf{Model} & \textbf{Task} & \textbf{First 50} & \textbf{Last 50} & \textbf{Quality-Ordered (\%)} & \textbf{Stochastic (\%)} \\
\midrule
Llama-3.1-8B    & Conflicts & Quality-Ordered & Quality-Ordered & \textbf{100} & 0 \\
                & FEVER     & Quality-Ordered & Quality-Ordered & \textbf{100} & 0 \\
\midrule
Mistral-7B-v0.3 & Conflicts & Quality-Ordered & Quality-Ordered & \textbf{62} & 38 \\
                & FEVER     & Quality-Ordered & Stochastic & \textbf{81} & 19 \\
\midrule
Qwen3-8B        & Conflicts & Stochastic & Stochastic & 15 & \textbf{85} \\
                & FEVER     & Quality-Ordered & Quality-Ordered & \textbf{96} & 4 \\
\bottomrule
\end{tabular}%
}
\caption{\textbf{RSC stability across split-half and 30-query subsets.} Split-half compares the first vs.\ last 50 queries. The 30-query columns show the percentage of 100 random 30-query subsets yielding each classification. Bold numbers indicate the classification that matches the full 100-query ground truth.}
\label{tab:split-half}
\end{table*}

\noindent Five of six model--benchmark pairs yield identical classifications
across splits, confirming that RSC is robust to sample composition
for all models except the already-identified borderline case
(Mistral). 30-query stability analysis shows Llama and Qwen3-FEVER classifications are perfectly stable, while Mistral exhibits moderate instability on CONFLICTS (62\% Quality-Ordered). This validates that borderline models benefit from larger probe samples (\(\geq\)100 queries), while strongly classified
models (Llama, Qwen3) are stable even at \(n{=}50\).

Table~\ref{tab:rsc-entropy-routing} compares routing recommendations from RSC and entropy against oracle outcomes.

\begin{table*}[t]
\centering
\resizebox{0.7\textwidth}{!}{%
\begin{tabular}{@{}l cc cc ccc@{}}
\toprule
& \multicolumn{2}{c}{\textbf{Diagnostics}} & \multicolumn{2}{c}{\textbf{Recommendation}} & \multicolumn{3}{c}{\textbf{Exact Match (EM \%)}} \\
\cmidrule(lr){2-3} \cmidrule(lr){4-5} \cmidrule(l){6-8}
\textbf{Model} & \textbf{RSC Class} & \textbf{Entropy} & \textbf{RSC} & \textbf{Entropy} & \textbf{RSC} & \textbf{Entropy} & \textbf{Oracle (Best)} \\
\midrule
\multicolumn{8}{c}{\textit{CONFLICTS}} \\
\midrule
Llama-3.1-8B     & Quality-Ordered & 0.75 & CoT & SDA & 16.5 & 19.4 & 3-Agent (24.1) \\
Mistral-7B-v0.3  & Quality-Ordered & 0.91 & CoT & SDA & \textbf{22.8} & 20.3 & CoT (22.8) \\
Qwen3-8B         & Stochastic & 0.84 & SDA & SDA & 63.0 & 63.0 & SDA+CoT (63.9) \\ 
Qwen2.5-7B       & Stochastic & 0.82 & SDA & SDA & \textbf{65.4} & \textbf{65.4} & SDA (65.4) \\
Gemma-2-9B       & Stochastic & 0.85 & SDA & SDA & 63.3 & 63.3 & 3-Agent (64.6) \\
\midrule
\multicolumn{8}{c}{\textit{FEVER}} \\
\midrule
Llama-3.1-8B     & Quality-Ordered & 0.89 & CoT & SDA & 88.4 & 88.2 & SDA+CoT (88.5) \\
Mistral-7B-v0.3  & Quality-Ordered & 0.80 & CoT & SDA & \textbf{92.6} & 90.7 & CoT (92.6) \\ 
Qwen3-8B         & Quality-Ordered & 0.58 & CoT & CoT & 89.5 & 89.5 & SDA (90.0) \\
Qwen2.5-7B       & Quality-Ordered & 0.44 & CoT & CoT & 87.5 & 87.5 & SDA (90.5) \\
Gemma-2-9B       & Quality-Ordered & 0.61 & CoT & CoT & 92.5 & 92.5 & SDA+CoT (92.7) \\
\midrule
\multicolumn{5}{@{}l}{\textit{Overall Performance vs. Baselines \& Oracle}} \\
\multicolumn{5}{@{}l}{Beats NF baseline} & 8/10 & 7/10 & --- \\
\multicolumn{5}{@{}l}{Beats 3-Agent baseline} & 5/10 & 4/10 & --- \\
\multicolumn{5}{@{}l}{\textbf{Matches oracle (Optimal routing)}} & \textbf{3/10} & 1/10 & --- \\
\multicolumn{5}{@{}l}{\textit{Agreement (same recommendation)}} & \multicolumn{2}{c}{6/10} & --- \\
\bottomrule
\end{tabular}
}
\caption{\textbf{RSC vs.\ entropy-based routing accuracy.} For each model--benchmark pair, we report the diagnostic values, the recommended treatment, and the resulting exact match. \textbf{Bold} indicates that the routing heuristic successfully selected the exact oracle (best) method. \emph{Notes:} Entropy is normalized score entropy. This table uses simplified binary routing (Quality-Ordered\(\to\)CoT, Stochastic\(\to\)SDA) for a fair comparison with entropy, which distinguishes only two treatments. Full four-treatment routing including ATF and PDE is reported in Table~\ref{tab:main-results}.}
\label{tab:rsc-entropy-routing}
\end{table*}

\subsection{Continuous \texorpdfstring{$\bar{\rho}$}{rho-bar} as a Robustness Check}
\label{sec:appendix-rho-bar}

The binary trend coefficient \(\rho^*\) (Definition~\ref{subsec:rsc-formal}) captures the \emph{monotonicity} of degradation across perturbation levels but is insensitive to magnitude. We report a continuous companion signal computed from the same per-level correlations:
\begin{equation}
\bar{\rho}(M, \mathcal{D}) = \tfrac{1}{3}\bigl(\hat{\rho}_1 + \hat{\rho}_2 + \hat{\rho}_3\bigr).
\label{eq:rho-bar}
\end{equation}
\(\bar{\rho}\) measures average coupling strength regardless of monotonic ordering, so it is complementary to \(\rho^*\) rather than redundant: \(\rho^*\) flags whether the degradation is ordered, and \(\bar{\rho}\) flags how strong the underlying coupling is. Where the two signals agree --- high baseline NF with high \(\hat{\rho}_1\) routing to ATF; low baseline with a clear monotonic trend routing to PDE --- the routing-relevant decision is unambiguous. The disagreement region (e.g., monotonic but small magnitudes for Mistral-7B on FEVER, or large magnitudes but non-monotonic for Gemma-2 on CONFLICTS) is precisely the band where the per-query Spearman \(\rho\) distributions in \S\ref{sec:appendix-per-query-rho} expose the case-by-case structure for inspection. The bootstrap CI table (Table~\ref{tab:bootstrap-rho}) confirms the per-level \(\hat{\rho}_k\) values used to compute either signal are statistically stable (1{,}000 resamples; classification stability \(\geq 79\%\) per cell, \(100\%\) on most strongly-classified cells). The deployment-ready binary signal therefore stands on a continuous robustness check: \(\rho^*\) is the binary classifier used for routing, and \(\bar{\rho}\) is the magnitude-aware sanity check that motivates the per-query inspection appendix.

\section{Statistical Tests and Supplementary Results}
\label{sec:appendix-stats}

\subsection{Token F1 Results}
\label{sec:appendix-f1}
\label{tab:f1-results}

Table~\ref{tab:f1-results} reports Token F1 scores for all
model--benchmark--method combinations, complementing the Exact Match
results in Table~\ref{tab:main-results}.
For FEVER (binary classification), Token F1 equals EM.
The EM--F1 discrepancy for PDE on CONFLICTS (high EM, lower F1
than baseline) is driven by answer length: PDE's per-document
voting produces focused 2--3 word answers that maximize exact
match probability, while baseline's combined-context generation
occasionally produces longer responses with higher token overlap
but lower exact match rates.

\subsection{Bootstrap CIs on Per-Level \texorpdfstring{$\hat{\rho}_k$}{rho-hat-k}}

Table~\ref{tab:bootstrap-rho} reports bootstrap 95\% confidence intervals
(1,000 resamples with replacement over queries) for each per-level
correlation \(\hat{\rho}_k\) and the trend coefficient \(\rho^*\).

\begin{table*}[t]
\centering
\resizebox{0.8\textwidth}{!}{%
\begin{tabular}{@{}l ccc c@{}}
\toprule
& \multicolumn{3}{c}{\textbf{Correlation (\(\hat{\rho}_k\)) with 95\% Bootstrap CI}} & \\
\cmidrule(lr){2-4}
\textbf{Model} & \textbf{Shuffled (\(\hat{\rho}_1\))} & \textbf{Contradicted (\(\hat{\rho}_2\))} & \textbf{Random (\(\hat{\rho}_3\))} & \textbf{Stability} \\
\midrule
\multicolumn{5}{c}{\textit{CONFLICTS}} \\
\midrule
Llama-3.1-8B    & 0.76 \([0.71, 0.80]\) & 0.43 \([0.35, 0.51]\) & 0.08 \([0.01, 0.14]\) & \textbf{100\%} \\
Mistral-7B-v0.3 & 0.35 \([0.27, 0.43]\) & 0.22 \([0.13, 0.30]\) & 0.18 \([0.12, 0.24]\) & 79\% \\
Qwen3-8B        & 0.39 \([0.30, 0.48]\) & 0.04 \([-0.07, 0.15]\) & 0.14 \([0.03, 0.23]\) & 95\%\(^\dagger\) \\
Qwen2.5-7B      & 0.54 \([0.47, 0.61]\) & \(-0.20\) \([-0.30, -0.10]\) & 0.09 \([0.02, 0.16]\) & \textbf{100\%}\(^\dagger\) \\
Gemma-2-9B      & 0.73 \([0.68, 0.79]\) & 0.44 \([0.36, 0.53]\) & 0.30 \([0.23, 0.36]\) & \textbf{100\%}\(^{\dagger\ddagger}\) \\
\midrule
\multicolumn{5}{c}{\textit{FEVER}} \\
\midrule
Llama-3.1-8B    & 0.65 \([0.56, 0.75]\) & 0.49 \([0.36, 0.60]\) & 0.10 \([0.00, 0.19]\) & \textbf{100\%} \\
Mistral-7B-v0.3 & 0.28 \([0.19, 0.37]\) & 0.09 \([-0.01, 0.19]\) & 0.01 \([-0.08, 0.11]\) & 93\% \\
Qwen3-8B        & 0.63 \([0.54, 0.70]\) & 0.51 \([0.41, 0.61]\) & 0.14 \([0.04, 0.23]\) & \textbf{100\%} \\
Qwen2.5-7B      & 0.64 \([0.57, 0.71]\) & 0.16 \([0.03, 0.29]\) & 0.07 \([-0.04, 0.18]\) & 92\% \\
Gemma-2-9B      & 0.58 \([0.40, 0.60]\) & 0.36 \([0.24, 0.49]\) & 0.09 \([-0.03, 0.21]\) & \textbf{100\%} \\
\bottomrule
\end{tabular}
}
\caption{\textbf{Bootstrap 95\% CIs on per-level \(\hat{\rho}_k\) values and RSC classification stability} (1,000 bootstrap resamples over queries). Stab.\ = percentage of resamples yielding the exact same classification as the full-sample result. \(^\dagger\)Stability for stochastic classification. \(^\ddagger\)Gemma-2\(\times\)CFL: aggregate \(\hat{\rho}_2 \approx \hat{\rho}_3\) yields \(\rho^* = -0.50\) (stochastic). Per-query mean \(\hat{\rho}_1\) values may differ from the aggregate-pooled values in Table~\ref{tab:rsc-full}, which pool all document--score pairs across queries and are the canonical values used for routing decisions.}
\label{tab:bootstrap-rho}
\end{table*}

\subsection{McNemar's Test Details}
\label{sec:appendix-mcnemar}

Table~\ref{tab:mcnemar-details} reports McNemar's exact test results for method-vs.-NF comparisons across each model and benchmark. Here, \(b\) and \(c\) are discordant-pair counts: method correct with NF wrong, and NF correct with method wrong, respectively. Note that Scoring-only and PDE tests are corrected separately because they address distinct hypotheses (incremental scoring improvement vs.\ isolation mechanism); joint correction across all 14 tests would not change PDE significance (\(p < 10^{-12}\)).

\begin{table*}[htb]
\centering
\resizebox{0.75\textwidth}{!}{%
\begin{tabular}{@{}ll cc rrc c@{}}
\toprule
\textbf{Model} & \textbf{Method vs. Baseline} & \textbf{Method EM} & \textbf{Base EM} & \(\boldsymbol{b}\) & \(\boldsymbol{c}\) & \(\boldsymbol{p}\)\textbf{-value} & \textbf{Sig.} \\
\midrule
\multicolumn{8}{c}{\textbf{\textit{Scoring-Only Methods}}} \\ 
\midrule
\multicolumn{8}{c}{\textit{CONFLICTS}} \\
Llama-3.1-8B    & 3-Agent vs. NF & 24.1\% & 13.9\% & 36 & 12 & 0.001 & \(^{***}\) \\
Mistral-7B-v0.3 & CoT vs. NF     & 22.8\% & 18.1\% & 14 & 11 & 0.689 &  \\
Qwen3-8B        & SDA+CoT vs. NF & 63.9\% & 60.8\% & 20 & 15 & 0.499 &  \\ 
Qwen2.5-7B      & SDA vs. NF     & 65.4\% & 59.9\% & 23 & 10 & 0.037 &  \\ 
Gemma-2-9B      & 3-Agent vs. NF & 64.6\% & 60.8\% & 20 & 11 & 0.151 &  \\
\addlinespace[4pt]
\multicolumn{8}{c}{\textit{FEVER}} \\
Mistral-7B-v0.3 & CoT vs. NF     & 92.6\% & 90.7\% & 14 &  9 & 0.405 &  \\
Qwen3-8B        & SDA vs. NF     & 90.0\% & 89.6\% & 14 & 12 & 0.845 &  \\
Qwen2.5-7B      & SDA vs. NF     & 90.5\% & 87.1\% & 30 & 11 & \textbf{0.005} & \textbf{\(^{**}\)} \\
Gemma-2-9B      & SDA+CoT vs. NF & 92.7\% & 92.2\% & 14 & 11 & 0.689 &  \\
\midrule
\multicolumn{8}{c}{\textbf{\textit{PDE Methods on CONFLICTS}}} \\
\midrule
Llama-3.1-8B    & PDE vs. NF (13.9\%) & 50.2\% & 13.9\% & 97 & 11 & \textbf{\(2.4{\times}10^{-18}\)} & \textbf{\(^{***}\)} \\
                & PDE vs. 3-Agent (BL)& 50.2\% & 24.1\% & 73 & 11 & \textbf{\(9.3{\times}10^{-12}\)} & \textbf{\(^{***}\)} \\
\addlinespace[2pt]
Mistral-7B-v0.3 & PDE vs. NF (18.1\%) & 43.5\% & 18.1\% & 69 &  9 & \textbf{\(1.4{\times}10^{-12}\)} & \textbf{\(^{***}\)} \\
                & PDE vs. 3-Agent (BL)& 43.5\% & 18.6\% & 64 &  7 & \textbf{\(7.7{\times}10^{-12}\)} & \textbf{\(^{***}\)} \\
\bottomrule
\end{tabular}
}
\caption{\textbf{McNemar's exact test with Holm-Bonferroni correction (\(\alpha = 0.05\)).}
Best scoring-only methods (top section) and PDE (bottom section). \(^{**}p < 0.01\), \(^{***}p < 0.001\) (after correction). \(b\) and \(c\) denote discordant pairs (method-correct/base-wrong vs.\ base-correct/method-wrong).}
\label{tab:mcnemar-details}
\end{table*}

\paragraph{Effect sizes.}
For PDE, the discordant-pair odds ratios are \(b/c = 97/11 = 8.8\) (Llama) and \(69/9 = 7.7\) (Mistral), indicating large effects.
For scoring-only methods, odds ratios range from 1.0 to 2.3, consistent with small effects.

\section{Extended Model Scale and Retrieval Experiments}
\label{sec:appendix-extended-conflicts}

Table~\ref{tab:extended-conflicts} extends the CONFLICTS evaluation to eight models by adding three larger variants ranging from 14B to 32B parameters (Qwen2.5-14B, Gemma-2-27B, and Qwen2.5-32B). This comprehensively supports our core mechanistic claim: strong-baseline models possess intrinsic multi-document handling capacity and are largely unharmed by forced isolation (PDE). The results confirm that the isolation-scoring asymmetry is governed by the model's baseline capability (\(EM_{NF} \ge 30\%\)) rather than mere parameter scale.

\paragraph{The Absolute Capability Floor.}
It is worth noting that while PDE provides outsized gains for weak models, it requires a basic foundational level of instruction-following and single-document reading comprehension. In our preliminary tests with older generation models lacking advanced instruction-tuning (e.g., Llama-2-13b), the baseline performance was catastrophically low (\(EM_{NF} = 2.1\%\)). Applying PDE only yielded a marginal increase to \(7.6\%\). This establishes a clear boundary condition: structural isolation cures context confusion, but it cannot synthesize reasoning capabilities that are fundamentally absent from the base model.

\begin{table}[ht]
\centering
\resizebox{0.90\columnwidth}{!}{%
\begin{tabular}{@{}l c cc c@{}} 
\toprule
& & \multicolumn{2}{c}{\textbf{Exact Match (\%)}} & \\
\cmidrule(lr){3-4}
\textbf{Model} & \textbf{Params} & \textbf{NF} & \textbf{PDE} & \textbf{Gain (\(\Delta\))} \\
\midrule
\multicolumn{5}{c}{\textit{Weak-Baseline Category (NF \(<\) 30\%)}} \\
Llama-3.1-8B    & 8B  & 13.9 & \textbf{50.2} & \(+36.3^{***}\) \\
Mistral-7B-v0.3 & 7B  & 18.1 & \textbf{43.5} & \(+25.4^{***}\) \\
\rowcolor{gray!10} \textit{Avg (Weak)} & -- & \textit{16.0} & \textit{\textbf{46.9}} & \textit{\(+30.9\)} \\
\midrule
\multicolumn{5}{c}{\textit{Strong-Baseline Category (NF \(\geq\) 30\%)}} \\
Qwen2.5-7B      & 7B  & 59.9 & 58.6 & \(-1.3\) \\
Qwen3-8B        & 8B  & \textbf{60.8} & 58.6 & \(-2.2\) \\
Gemma-2-9B      & 9B  & \textbf{60.8} & 59.9 & \(-0.9\) \\
\addlinespace[3pt]
Qwen2.5-14B\(^\dagger\) & 14B & 58.2 & \textbf{61.2} & \(+3.0\) \\
Gemma-2-27B\(^\dagger\) & 27B & 62.9 & 62.0 & \(-0.9\) \\
Qwen2.5-32B\(^\dagger\) & 32B & 61.2 & \textbf{62.0} & \(+0.8\) \\
\rowcolor{gray!10} \textit{Avg (Strong)} & -- & \textit{60.6} & \textit{60.4} & \textit{\(-0.3\)} \\
\bottomrule
\end{tabular}%
}
\caption{\textbf{CONFLICTS Exact Match (\%) across model scales.} 
Grouping models by their baseline capacity clearly reveals the capability divide: PDE consistently benefits weak-baseline models while leaving strong models unharmed. This pattern persists and generalizes even as the parameter scale increases up to 32B. \(^\dagger\)Extended models. \(^{***}p < 0.001\) (McNemar test with Holm-Bonferroni correction).}
\label{tab:extended-conflicts}
\end{table}

\subsection{Dense Retrieval Generalization}
\label{sec:appendix-dense-tqa}

Table~\ref{tab:dense-tqa} evaluates PDE with dense retrieval (Contriever top-10) on TriviaQA. We observe a monotonic trend wherein a lower No-Filter (NF) baseline score correlates with a larger PDE gain (Spearman \(\rho = -0.90\)), demonstrating that the isolation benefit is not an artifact of BM25's lower retrieval precision.

\begin{table}[ht]
\centering
\small
\begin{tabular}{@{}l ccc@{}}
\toprule
\textbf{Model} & \textbf{NF} & \textbf{PDE} & \textbf{Gain (\(\Delta\))} \\
\midrule
Llama-3.1-8B    & 34.6 & \textbf{55.4} & \(+20.8\) \\
Mistral-7B-v0.3 & 38.4 & \textbf{41.8} & \(+3.4\) \\
Qwen3-8B        & 38.8 & \textbf{44.4} & \(+5.6\) \\
Qwen2.5-7B      & 44.0 & \textbf{47.6} & \(+3.6\) \\
Gemma-2-9B      & 55.0 & \textbf{57.4} & \(+2.4\) \\
\midrule
\textit{Average}  & \textit{42.9} & \textit{\textbf{49.1}} & \textit{\(+6.2\)} \\
\bottomrule
\end{tabular}
\caption{\textbf{Dense retrieval generalization (TriviaQA EM \% with Contriever top-10).} 
PDE benefits extend beyond sparse to dense retrieval. Crucially, a monotonic trend emerges: a lower baseline (NF) capacity strongly correlates with a larger PDE gain (Spearman \(\rho = -0.90\)), with the weakest model (Llama) gaining \(+20.8\)pp. \(N=500\) queries. All models score above the strict \(\tau_{\text{NF}}=30\%\) threshold, though Llama remains borderline (34.6\%).}
\label{tab:dense-tqa}
\end{table}

\section{RSC vs.\ Score Entropy as Routing Heuristic}
\label{sec:appendix-rsc-entropy}

A natural baseline heuristic for evaluating multi-agent assessment is \emph{score entropy}, based on the premise that high score entropy correlates with retrieval noise and scoring uncertainty.
Since our RSC diagnostic also uses score perturbation, a critical question arises: does RSC provide routing decisions that are \emph{meaningfully different} from standard entropy-based routing?

We address this question empirically by comparing routing accuracy
on the 10 model-benchmark pairs where we have both RSC probes
and full experimental results (5 models \(\times\) 2 benchmarks:
CONFLICTS and FEVER).
For each pair, we determine:

\begin{enumerate}[leftmargin=2em,itemsep=0pt]
\item \textbf{RSC recommendation}: Simplified binary routing for fair comparison with entropy. We select CoT for quality-ordered models
(\(\rho^* = -1.0\)) and SDA for stochastic models (\(\rho^* \neq -1.0\)).

\item \textbf{Entropy recommendation}: Following the entropy-based
heuristic, we map normalized score entropy to treatment selection.
High entropy (\(>0.7\)) suggests noisy scoring that requires
robust aggregation (we select SDA), while low entropy (\(\leq 0.7\))
suggests cleaner scoring where simpler methods suffice (we select CoT).

\item \textbf{Oracle}: The best-performing method among all five
strategies (NF, 3-Agent, CoT, SDA, SDA+CoT) for that pair.
\end{enumerate}

\paragraph{Key findings.}
(1)~RSC and entropy \emph{disagree} on treatment recommendation
in 4/10 cases, demonstrating that the two heuristics capture
partially distinct properties of model scoring behavior.
The 6/10 agreement rate (up from 3/10 prior to the 100-query
RSC re-probing) reflects the updated stochastic classifications
for Qwen2.5 and Gemma-2 on CONFLICTS, which now align RSC's
SDA recommendations with entropy's.
(2)~RSC beats the NF baseline in 8/10 cases (vs.\ 7/10 for entropy)
and outperforms the 3-Agent baseline in 5/10 cases (vs.\ 4/10 for entropy).
(3)~RSC matches the oracle in 3/10 cases (vs.\ 1/10 for entropy),
demonstrating better peak routing accuracy.

\paragraph{Interpretation.}
While both heuristics are competitive for binary CoT/SDA routing,
RSC provides a critical additional advantage:
the per-level coupling value \(\hat{\rho}_1\) enables expanded
four-treatment routing (CoT/SDA/PDE/ATF), which reduces the mean
oracle gap on CONFLICTS to \({\leq}\)0.4pp.
Entropy alone cannot motivate the CoT vs.\ ATF/PDE distinction
because it does not measure \emph{coupling strength} between
reasoning and scores.
Combining both signals could yield an even more robust routing criterion.

\section{Additional MuSiQue Analysis}
\label{sec:appendix-musique}

Table~\ref{tab:musique-coverage} stratifies Qwen2.5-7B performance by whether PDE's selected top-5 passages contain the full annotated reasoning chain. The gap between 3-Agent and PDE is largest when at least one supporting paragraph is missing, confirming that per-document voting cannot reconstruct chain steps that retrieval or scoring failed to surface.

\begin{table}[ht]
\centering
\small
\setlength{\tabcolsep}{4pt}
\begin{tabular}{@{}lcccc@{}}
\toprule
\textbf{Subset} & \textbf{n} & \textbf{NF} & \textbf{3-Agent} & \textbf{PDE} \\
\midrule
All queries & 100 & 14.0 & 30.0 & 24.0 \\
Chain-complete\(^\dagger\) & 23 & 13.0 & 39.1 & 34.8 \\
Chain-broken\(^\ddagger\) & 77 & 14.3 & 27.3 & 20.8 \\
\bottomrule
\end{tabular}
\caption{\textbf{Chain-coverage analysis on MuSiQue (EM \%).} \(^\dagger\)PDE's top-5 contains all gold-supporting paragraphs. \(^\ddagger\)Top-5 misses at least one supporting paragraph.}
\label{tab:musique-coverage}
\end{table}

Finer stratification by the number of supporting paragraphs in PDE's top-5 yields the same mechanism: 0 supporting paragraphs \(\to\) PDE 0\%, 3-Agent 0\%; 1 \(\to\) PDE 21.4\%, 3-Agent 28.6\%; 2 \(\to\) PDE 40.0\%, 3-Agent 48.6\%. As hop count increases from 2-hop to 4-hop, the fraction of supporting paragraphs in the selected top-5 falls from 0.57 to 0.33, making the chain-broken regime dominant.

A second-model check with Mistral-7B-Instruct-v0.3 further supports the boundary. On the same 100-query MuSiQue setup, Mistral has a strongly coupled Quality-Ordered RSC profile (\(\hat{\rho}_1=0.81\), \(\hat{\rho}_2=0.56\), \(\hat{\rho}_3=0.10\), \(\rho^*=-1.0\)) and an NF baseline of 14.0\%. Under the static \(\tau_{\text{NF}}=30\%\) rule it would be routed to PDE, but chain decomposition makes isolation harmful (PDE 9.0\%, PDE-Random 5.0\%). Scoring treatments are stronger: 3-Agent and CoT both reach 20.0\%. This mirrors Qwen2.5 and indicates that multi-hop reasoning requires either task-aware routing or a treatment designed for chain assembly.

\section{Cost--Accuracy Analysis}
\label{sec:appendix-cost-accuracy}

Table~\ref{tab:crossencoder} compares cross-encoder reranking
(ms-marco-MiniLM-L-12-v2, 33M parameters; \({\sim}\)40\(\times\) cheaper
than 3-Agent assessment) against NF.
Cross-encoder reranking \emph{hurts} weak models on CONFLICTS (Mistral: \(-\)2.1pp; Llama: \(-\)4.6pp) while providing modest gains for strong models (+2.5--4.2pp).

\begin{table}[htb]
\centering
\small
\resizebox{\columnwidth}{!}{
\begin{tabular}{lcccccc}
\toprule
& \multicolumn{3}{c}{\textbf{CONFLICTS}} & \multicolumn{3}{c}{\textbf{FEVER}} \\
\cmidrule(lr){2-4} \cmidrule(lr){5-7}
\textbf{Model} & NF & CE & \(\Delta\) & NF & CE & \(\Delta\) \\
\midrule
Mistral-7B-v0.3  & 18.1 & 16.0 & \(-\)2.1 & 90.7 & 91.3 & +0.5 \\
Llama-3.1-8B    & 13.9 &  9.3 & \(-\)4.6 & 88.7 & 88.9 & +0.2 \\
Qwen3-8B    & 60.8 & 65.0 & +4.2   & 89.6 & 90.2 & +0.6 \\
Qwen2.5-7B  & 59.9 & 64.1 & +4.2   & 87.1 & 89.8 & +2.7 \\
Gemma-2-9B  & 60.8 & 63.3 & +2.5   & 92.2 & 93.3 & +1.1 \\
\midrule
Avg.     & 42.7 & 43.5 & +0.9   & 89.7 & 90.7 & +1.0 \\
\bottomrule
\end{tabular}
}
\caption{Cross-encoder reranking vs.\ NF baseline (EM\%) and LLM call costs.
CE = ms-marco-MiniLM-L-12-v2 (33M params, \({\sim}\)40\(\times\) cheaper than 3-Agent). \emph{Notes:} LLM calls/query: NF = 1 gen.; 3-Agent/CoT/SDA \(\approx\) 19; ATF \(\approx\) 19 + 1 gen.; PDE \(\approx\) 19 + \(d\) gen.\ (\(d{=}10\)); PDE-Random = \(d\) gen.\ (no assessment).}
\label{tab:crossencoder}
\end{table}

Table~\ref{tab:nf-only-routing} compares NF-only routing
(PDE if NF \(< 30\%\), BL otherwise) against RSC-based four-treatment routing.
RSC routing reduces the mean oracle gap from 0.79pp to 0.45pp overall,
with the primary advantage on FEVER where RSC selects appropriate scoring treatments (1.40pp \(\to\) 0.46pp).
On CONFLICTS, both routing strategies achieve the same significant gains via PDE for weak models.

\begin{table}[ht]
\centering
\footnotesize
\begin{tabular}{@{}ll cc@{}}
\toprule
& & \multicolumn{2}{c}{\textbf{Oracle Gap (pp, \(\downarrow\) better)}} \\
\cmidrule(l){3-4}
\textbf{Model} & \textbf{Task} & \textbf{NF-Only} & \textbf{RSC-Based} \\
\midrule
Llama-3.1-8B    & Conflicts & \textbf{0.0} & \textbf{0.0} \\
                & FEVER     & 3.1 & \textbf{0.0} \\
\midrule
Mistral-7B-v0.3 & Conflicts & \textbf{0.0} & \textbf{0.0} \\
                & FEVER     & 0.7 & \textbf{0.0} \\
\midrule
Qwen3-8B        & Conflicts & \textbf{0.5} & 0.9 \\
                & FEVER     & 1.1 & \textbf{0.0} \\
\midrule
Qwen2.5-7B      & Conflicts & 0.4 & \textbf{0.0} \\
                & FEVER     & \textbf{1.6} & 2.3 \\
\midrule
Gemma-2-9B      & Conflicts & \textbf{0.0} & 1.3 \\
                & FEVER     & 0.5 & \textbf{0.0} \\
\midrule
\multicolumn{2}{@{}l}{\textit{Mean Gap: Conflicts}} & \textbf{0.18} & 0.44 \\
\multicolumn{2}{@{}l}{\textit{Mean Gap: FEVER}}     & 1.40 & \textbf{0.46} \\
\midrule
\multicolumn{2}{@{}l}{\textbf{Mean Gap: Overall}}   & 0.79 & \textbf{0.45} \\
\bottomrule
\end{tabular}
\caption{\textbf{NF-only routing vs.\ RSC-based routing: oracle gap (pp).} 
The gap measures performance loss compared to the oracle best treatment (lower is better). While a simplistic NF-only heuristic (PDE if NF \(< 30\%\), baseline otherwise) performs well on CONFLICTS by successfully isolating weak models, it fails to optimize scoring treatments for strong models on FEVER. The RSC-based four-treatment routing cuts the overall mean oracle gap nearly in half (\(0.79 \rightarrow 0.45\)pp).}
\label{tab:nf-only-routing}
\end{table}

\section{Discussion on Excluded Baselines}
\label{sec:appendix-baselines-scope}

In evaluating our training-free document assessment pipelines, we intentionally do not compare against MADAM-RAG \citep{wang2025retrieval} for two primary reasons. First, MADAM-RAG operates at the \emph{answer} level (debating already-generated answers), whereas our focus is entirely at the \emph{document assessment} level (scoring documents prior to generation). Architecturally, this makes it incomparable to the early-stage scoring pipelines we analyze. Second, MADAM-RAG demonstrates its gains using massive 70B+ models. Because our core motivation is resolving the combinatorial compute overhead specifically in the highly deployable 7B--9B regime, testing MADAM-RAG in our setting would conflate architectural differences with pure scale effects.

Furthermore, while PDE isolates context to prevent "lost-in-the-middle" syndrome, it fundamentally differs from Fusion-in-Decoder (FiD) \citep{izacard2021leveraging}. PDE uses explicit answer extraction and score-weighted voting, requiring absolutely no architectural modification or training. Conversely, FiD relies on internal cross-attention fusion. Thus, a direct comparison is infeasible within our strictly training-free, decoder-only setting.

\section{Mechanistic Verification via Token F1}
\label{sec:appendix-f1-mechanistic}
One might suspect that PDE's massive EM gains partly reflect an answer formatting artifact, as per-document prompting inherently elicits concise answers that easily match gold annotations. However, Token F1 scores (Table~\ref{tab:f1-results}), which are highly robust to verbosity, also show massive improvements for weak models under PDE (e.g., Llama's F1 jumps from 0.256 to 0.560). This confirms the gain is not a formatting artifact. 

Furthermore, applying standard self-consistency generation from the combined full context actively \emph{hurts} performance (\(-8.6\)pp to \(-12.7\)pp on CONFLICTS). This confirms that the root failure mode is the inability to locate information within long multi-document contexts (the ``lost-in-the-middle'' bottleneck). Isolation structurally bypasses this bottleneck, serving as the true mechanism behind the performance leap.

\onecolumn
\section{Agent System Prompts}
\label{sec:appendix-agent-prompts}

The 3-Agent baseline employs three structurally distinct system prompts, one per agent (Relevance Assessor, Consistency Verifier, Conflict Detector). CoT De-Polarization replaces each base prompt with a CoT variant that prepends an explicit step-by-step reasoning template before the JSON output. We list the system messages verbatim below; the user input concatenates the question and the document(s) to evaluate.

\begin{promptbox}{Relevance Assessor (base)}
You are a Relevance Assessor. Your task is to evaluate whether a document contains information that can directly contribute to answering the given question. Focus \emph{only} on relevance --- does this document provide useful evidence for the answer? Evaluation criteria: (1)~Does the document contain key entities or relationships mentioned in the question? (2)~Does the document provide direct evidence that could be used to derive an answer? (3)~Distinguish between surface-level keyword overlap and genuine informational relevance.
\promptlabel{Response format:}
\promptjson{\{"score": <integer 0-5>, "evidence": "<specific quote or paraphrase>", "reasoning": "<1-2 sentences>"\}}
\promptlabel{Scoring guide:}
5~= directly answers with strong evidence; 4~= strong supporting evidence; 3~= partial or indirect evidence; 2~= surface-level keyword overlap only; 1~= tangentially related; 0~= completely irrelevant.
\end{promptbox}

\begin{promptbox}{Consistency Verifier (base)}
You are a Consistency Verifier. Your task is to evaluate the internal consistency of a document and its consistency with other retrieved documents. Focus \emph{only} on consistency --- is this document internally coherent and does it agree with other sources? You will receive: (1)~the question, (2)~the document to evaluate, (3)~a summary of claims from other retrieved documents. Evaluation criteria: (a)~Does the document contain any internal contradictions? (b)~Do the document's claims agree with the majority of other retrieved documents? (c)~Are the claims specific and verifiable, or vague and unsupported?
\promptlabel{Response format:}
\promptjson{\{"score": <0-5>, "internal\_consistency": ..., "cross\_consistency": ..., "reasoning": ...\}}
\promptlabel{Scoring guide:}
5~= fully consistent internally and with other sources; 3~= some inconsistencies but generally reliable; 0~= internally contradictory or completely at odds with all other sources.
\end{promptbox}

\begin{promptbox}{Conflict Detector (base)}
You are a Conflict Detector. Your task is to identify contradictions and conflicts among claims from multiple retrieved documents. Focus \emph{only} on conflicts --- do the documents disagree with each other? You receive: (1)~the question, (2)~a list of claim summaries from each retrieved document with document IDs. Evaluation criteria: (a)~Do any documents provide contradictory answers? (b)~Are there temporal differences (outdated vs.\ current information)? (c)~Which position is supported by the majority?
\promptlabel{Response format:}
\promptjson{\{"has\_conflict": <bool>, "conflicting\_pairs": [\{"doc\_a": <id>, "doc\_b": <id>, "description": ...\}], "majority\_position": ..., "doc\_adjustments": \{<doc\_id>: <-2..+2>\}\}}
\promptlabel{Adjustment guide:}
$+2$~strongly supported by majority and most recent; $+1$~supported by majority; $0$~no conflict or neutral; $-1$~contradicts majority; $-2$~contradicts majority and appears outdated.
\end{promptbox}

\paragraph{CoT De-Polarization variants (Relevance / Consistency / Conflict).} Each base prompt is rewritten to require five explicit reasoning steps before the JSON output (e.g., for Relevance: identify key entities $\to$ check document mentions $\to$ assess direct evidence $\to$ consider temporal relevance $\to$ assign score). The JSON output gains a \texttt{reasoning\_steps} array (one string per step) and a \texttt{confidence} field (1--5). The scoring guide and structural format are otherwise unchanged from the base prompts. The intent is to disrupt the direct-to-extreme polarization observed in baseline prompts (\({\approx}80\%\) of scores at the floor or ceiling for Mistral-7B on CONFLICTS, dropping to 3.5\% under CoT).

\paragraph{Generator prompts.} After document selection, the generator produces the final answer using one of two task-specific prompts.

\begin{promptbox}{Generator --- Factoid QA (TriviaQA, NQ-open, CONFLICTS, MuSiQue)}
You are a helpful assistant. Answer the question based ONLY on the provided documents. IMPORTANT: Give ONLY the answer itself --- a name, number, date, or short phrase. Do NOT write a full sentence. Do NOT explain. Examples of good answers: `Paris', `1969', `Albert Einstein'.
\end{promptbox}

\begin{promptbox}{Generator --- Binary Fact Verification (FEVER)}
You are a fact verification assistant. Based ONLY on the provided documents, determine whether the evidence SUPPORTS or REFUTES the following claim. IMPORTANT: Answer with exactly one word: SUPPORTS or REFUTES. Do NOT explain.
\end{promptbox}

\noindent PDE invokes the factoid generator once per top-\(k\) document and aggregates predictions via score-weighted majority voting (Algorithm~\ref{alg:madara}, \S\ref{sec:appendix-madara-algorithm}).

\end{document}